%% file: main.tex
\documentclass[10pt,twocolumn,letterpaper]{article}

\usepackage[pagenumbers]{iccv} 


\definecolor{iccvblue}{rgb}{0.21,0.49,0.74}
\usepackage[pagebackref,breaklinks,colorlinks,allcolors=iccvblue]{hyperref}

\def\paperID{1751} 
\def\confName{ICCV}
\def\confYear{2025}

\usepackage{color}
\usepackage{amsmath}

\usepackage[utf8]{inputenc} 
\usepackage[T1]{fontenc}    
\usepackage{hyperref}       
\usepackage{url}            
\usepackage{booktabs}       
\usepackage{amsfonts}       
\usepackage{nicefrac}       
\usepackage{microtype}      
\usepackage{xcolor}         

\usepackage{graphicx}       
\usepackage{kotex}          
\usepackage{wrapfig}        
\usepackage{adjustbox}      
\usepackage{array,multirow} 
\usepackage{tcolorbox}
\usepackage{tikz}
\usepackage{duckuments}
\usepackage{ctable}
\usepackage{colortbl}

\usepackage{subcaption}
\DeclareMathOperator*{\argmax}{arg\,max}

\usepackage{amssymb}
\usepackage{times}
\usepackage{epsfig}
\usepackage{graphbox}
\usepackage{makecell}
\usepackage{xspace}
\usepackage{hyperref}

\title{Eye-for-an-eye: Appearance Transfer with Dense Semantic Correspondence in Diffusion Models}

\author{
Sooyeon Go \quad Kyungmook Choi \quad Minjung Shin \quad Youngjung Uh\thanks{Corresponding author}\\[3pt]
Yonsei University, Seoul, South Korea\\[3pt]
{\tt\small \{sooyeon8658, kyungmook.choi, smj139052, yj.uh\}@yonsei.ac.kr}
}

\input{macro.tex}

\begin{document}

\twocolumn[{%
\renewcommand\twocolumn[1][]{#1}%
\maketitle
\input{sources/teaser}
}]

\maketitle
\input{sec/0_abstract}    
\input{sec/1_intro}
\input{sec/2_rel_work}
\input{sec/3_method}
\input{sec/4_experiments}
\input{sec/5_conclusion}

\clearpage
{
    \small
    \bibliographystyle{ieeenat_fullname}
    \bibliography{cite}
}
\clearpage
\input{sec/6_appendix}

\end{document}

%% file: macro.tex
\newcommand{\fref}[1]{Fig. \ref{#1}}

\newcommand{\tref}[1]{Tab. \ref{#1}}
\newcommand{\aref}[1]{Appendix \ref{#1}}

\newcommand{\secref}[1]{Sec.~\ref{#1}}

\newcommand{\myparagraph}[1]{\smallskip\noindent{\textbf{#1} {} {}}}

\usepackage{diagbox}

%% file: sources/teaser.tex
\noindent
\makebox[\textwidth][c]{
    \includegraphics[width=\textwidth]{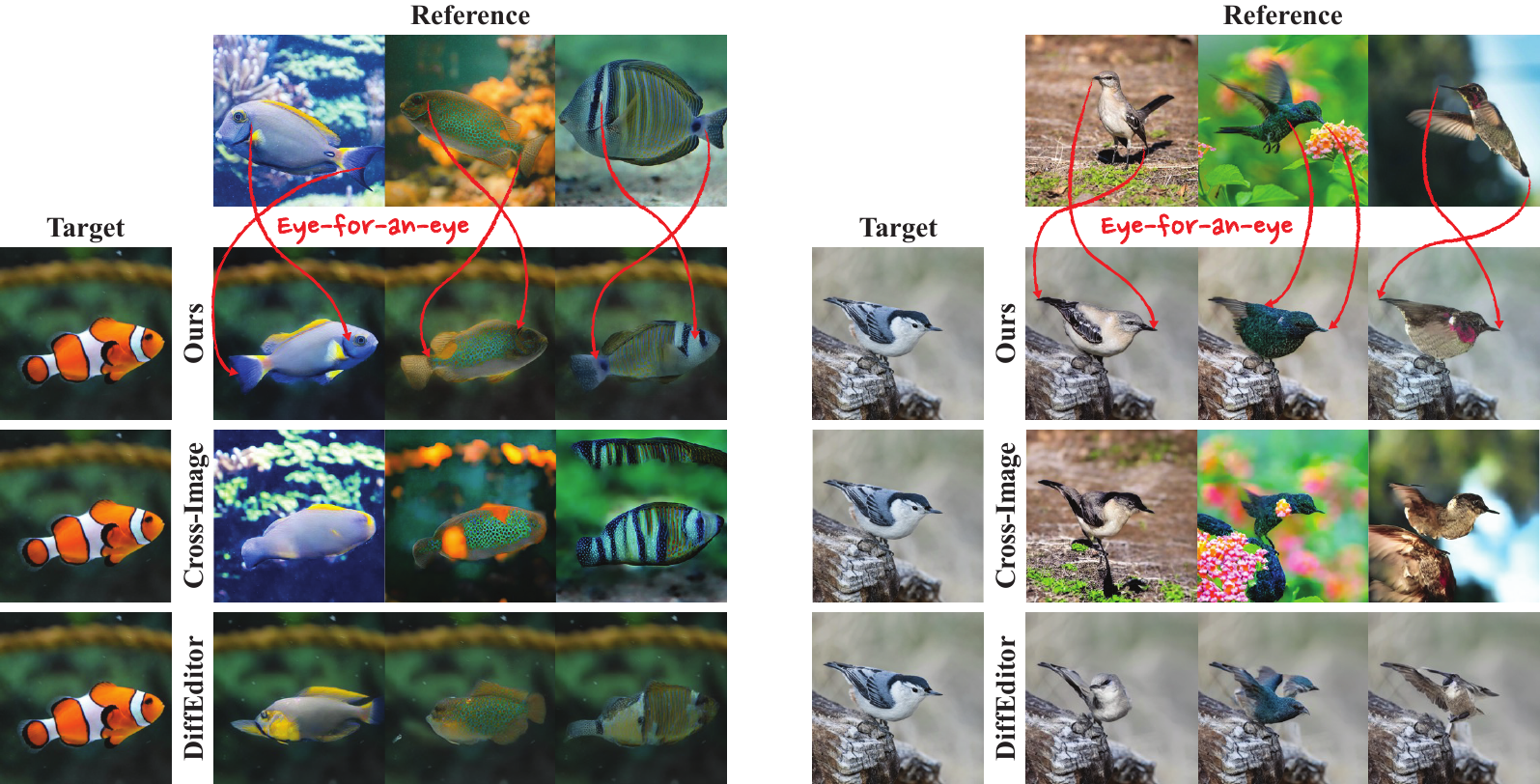}
}
\vspace{-5mm}
\begin{center}
    \captionof{figure}{Our method transfers semantically corresponding appearances from reference images to target images. In contrast to other methods such as DiffEditor \cite{diffeditor} and Cross-Image \cite{cross_image}, our method preserves the structure of the target images successfully transfers the colors and patterns considering the semantic meanings from the references.}
    \label{fig:teaser}
    \vspace{-3pt}
\end{center}

%% file: sec/0_abstract.tex
\begin{abstract}
As pre-trained text-to-image diffusion models have become a useful tool for image synthesis, people want to specify the results in various ways. This paper tackles training-free appearance transfer, which produces an image with the structure of a target image from the appearance of a reference image. 
Existing methods usually do not reflect semantic correspondence, as they rely on query-key similarity within the self-attention layer to establish correspondences between images.
To this end, we propose explicitly rearranging the features according to the \textbf{dense} semantic correspondences. Extensive experiments show the superiority of our method in various aspects: preserving the structure of the target and reflecting the correct color from the reference, even when the two images are not aligned.  
\end{abstract}

%% file: sec/1_intro.tex
\section{Introduction}
\label{sec:intro}

Text-to-image diffusion models \cite{stable} generate high-quality, realistic images from textual inputs. Although it allows users to easily describe the desired results, it falls short in more specific controls that are difficult to be described in texts. 
Alternatively, it is easier for the users to provide reference images and carrying their specific elements to the results.
Such elements include shapes \cite{controlnet}, main subject \cite{hertz2022prompt, ip2p}, and most of the images for partial editing \cite{blended,paint,diffedit}.
We tackle a scenario with two input images, where the result has the shape of one image and the color pattern of the other. It is often called appearance transfer from a reference image to a target image.

Although previous methods for appearance transfer \cite{cross_image,diffeditor,dragon,self_guidance} are promising, they struggle when the poses are not aligned. \fref{fig:teaser} shows that they often transfer eyes to tails and tails to heads. Hence, we hypothesize that the solution lies in establishing correspondences between the target and reference.
A straightforward solution for the above problem would be a two-stage procedure: finding semantic correspondences \cite{zhang2024tale, dift, telling} and following them to transfer the reference to the target. 
However, most semantic matches produce sparse key-point correspondences and dense correspondences are not accurate enough. Moreover, the two-stage approach is inherently cumbersome and costly.

In this paper, we analyze the limitations of previous methods with a self-attention injection or two-stage approach and propose \textit{Eye-for-an-eye}. It consists of three parts: finding correspondences, transferring features, and recursively running them through a generative process. As a whole, it considers \textit{dense semantic correspondences} to transfer the appearance of a reference image to the target image. Our method has non-trivial design choices as follows. We find that the cosine similarity between features of reference and target features before the self-attention layer allows for more semantically meaningful matching than the attention mechanism between the reference key and target query within self-attention. Then we replace the target features with the reference features rearranged according to our correspondence. It accurately keeps the target structure in the result. We recursively run this operation along the generative process. 

Our method accurately transfers the appearance from precise locations in a reference image even in challenging scenarios involving complex colors and patterns, or diverse views and poses. In addition, our results maintain the structure of the target image. We demonstrate the superiority of our method compared to previous methods with extensive qualitative and quantitative evaluation. Beyond intra-domain appearance transfer, our method generalizes to cross-domain appearance transfer and supports applying different appearances to multiple objects. Ours not only achieves superior appearance transfer results but also shows the best dense correspondence performance compared to existing semantic matching methods.

%% file: sec/2_rel_work.tex
\section{Related Work}

\paragraph{Appearance transfer}
Appearance transfer produces an image that combines the shape and color patterns of two different images.
This is accomplished by training on each target domain \cite{goel2023pair, chen2023anydoor, swap_ae} and using either input image pairs \cite{spliceVIT} or using external models to guide diffusion model \cite{diffuseit}.
While these methods maintain the structure of the target image, they tend to struggle with unaligned images or those from different domains.
Recent approaches \cite{self_guidance, dragon, diffeditor, cross_image} excel with images from different domains without requiring fine-tuning.
However, self-guidance \cite{self_guidance} leads to discrepancies in color distribution between the output and reference images 
because they make the \textit{average} features of the output similar to the reference.
The methods with key-value injection \cite{cross_image,dragon} expect the attention mechanism to find the semantic similarity for the transfer.
The attention mechanism often produces wrong semantic correspondences such as beaks to wings and tails to heads as shown in \cref{fig:teaser}.
In contrast, our method transfers appearance following correct semantic correspondence, even in a training-free manner. 

\begin{figure*}[t] \centering
    \includegraphics[width=\linewidth]{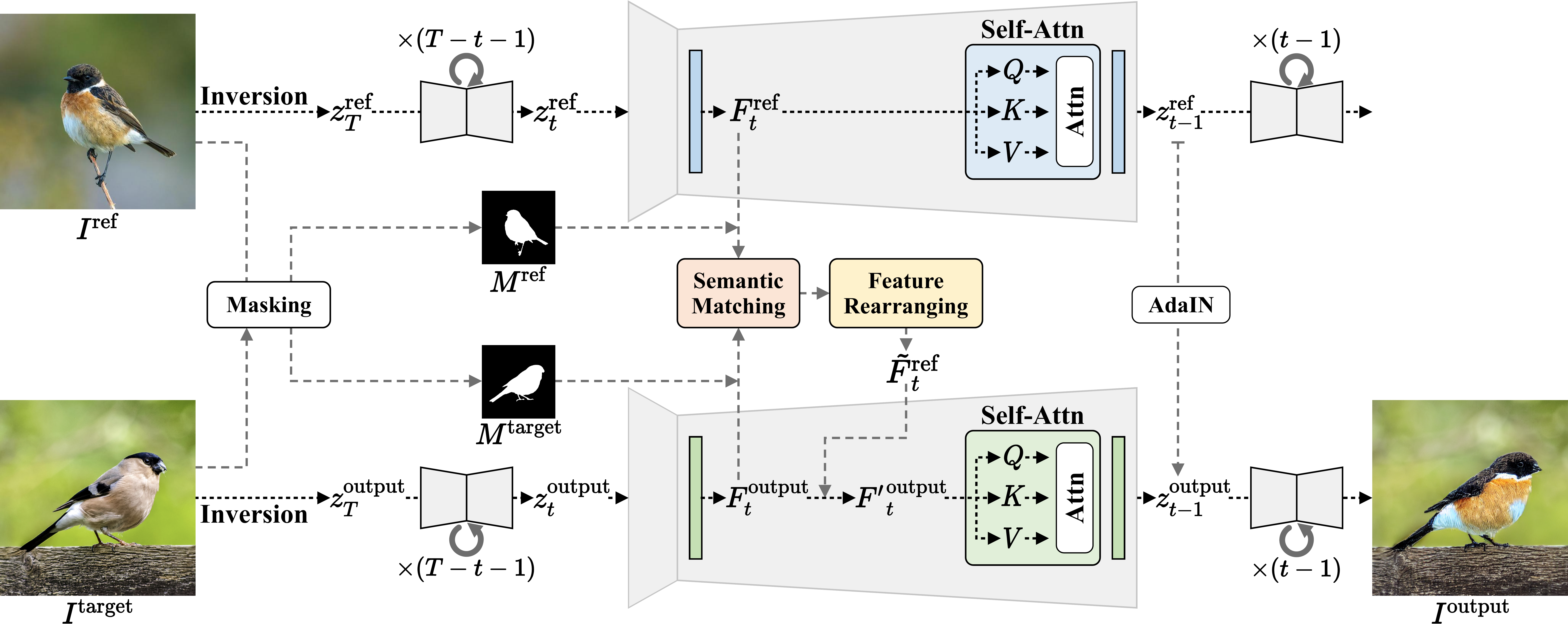}
    \caption{
        \textbf{Pipeline of our method.}
        We transfer the semantically corresponding appearance of objects from a reference image to a target image.
        Given $I^\text{ref}$, $I^\text{target}$, and their masks $M^\text{ref}$ and $M^\text{target}$, we find semantic correspondences between their features before the self-attention layers $F^\text{ref}_t$ and $F^\text{output}_t$.
        Then, we inject the rearranged features based on these correspondences.
    }
    \label{fig:overview}
    \vspace{-1.5mm}
\end{figure*}

\paragraph{Manipulating features for image editing}
Recent approaches \cite{hertz2022prompt, gu2024photoswap, parmar2023zero, lee2024conditional, tumanyan2023plug, masactrl, cross_image, lu2023tf, patashnik2023localizing, du2020energy} explore the manipulation of attention layers of pretrained diffusion models for image editing. 
In this context, PnP-diffusion \cite{tumanyan2023plug} leverages the semantic information in self-attention layers, demonstrating that modifying attention features can be used for editing tasks without requiring fine-tuning.
MasaCtrl \cite{masactrl} and Cross-Image \cite{cross_image} replace the key and value features in the self-attention layer to achieve text-guided translation of reference images.
However, we observe that query-key attention maps and the weighted summation in the self-attention are insufficient for transferring semantically matched appearances.
Therefore, instead of directly injecting entire key-value pairs or whole features, we propose injecting features after rearranging them based on their semantic correspondence.

\paragraph{Semantic correspondence}
Leveraging the diffusion features of models, unsupervised semantic correspondence methods \cite{hedlin2024unsupervised, dift} outperform other weakly-supervised methods.
SD-DINO \cite{zhang2024tale} further enhances this performance by incorporating DINO ViT \cite{amir2021deep} as an additional feature extractor.
Recent approaches \cite{tumanyan2023plug, freedom, masactrl} observe that semantic understanding in diffusion models is distributed across timesteps and U-Net layers.
Consequently, Diffusion Hyperfeatures \cite{luo2024diffusion} leverages these distributed representations by integrating feature maps across timesteps, demonstrating their effectiveness in keypoint correspondence tasks.
To incorporate semantic information distributed across timesteps into appearance transfer, we rearrange the feature maps according to the correspondences found at each timestep.

%% file: sec/3_method.tex
\section{Method}

We aim to transfer the appearance of objects from a reference image $I^\text{ref}$ to a target image $I^\text{target}$ based on semantic correspondences between them. 
The appearance includes attributes such as the color and pattern of the object.
As shown in \fref{fig:overview}, our method produces an image from an output denoising process (the target process being injected with reference features) starting from an inversion~\cite{edit_friendly} of $I^\text{target}$ (bottom) with modifications from another reference-denoising process starting from an inversion of $I^\text{ref}$ (top).
The modification includes finding semantic correspondences between the two denoising processes and injecting features with rearrangement.

\subsection{Revisiting of self-attention} \label{sec:motive}
\input{sources/motivation}

In the self-attention layer of the U-Net in Stable Diffusion \cite{stable}, an attention map is generated by applying a dot product and softmax to the query $Q$ and key $K$, which indicates positional similarities.
The weighted sum of this attention map and value $V$ allows each position in $Q$ to aggregate relevant information from similar positions in the $K$-$V$ pairs and $Q$ .

Recent appearance transfer methods \cite{cross_image,dragon} introduces $KV$ injection, which integrates $K_\text{ref}$-$V_\text{ref}$ pairs from a reference denoising process into the target denoising process.
During this process, the $KV$ injection aggregates $V_\text{ref}$ based on the attention map between $Q_\text{target}$ and $K_\text{ref}$.
Therefore, while the attention map of $Q_\text{target}$ and $K_\text{ref}$ indicates the similarity with $Q_\text{target}$ for determining the location from which to aggregate $V_\text{ref}$, it does not represent correspondence matching between the target image and the reference image, as shown in \fref{fig:motivation}.
As a result, this can lead to transfers with mismatched semantic meanings. 

Moreover, the self-attention with the KV injection aggregates features from \textit{multiple locations} of the reference rather than borrowing a feature from a single point. Although it might be a good way to transfer global style, it prevents the results from having clear local appearances.

In the following subsection, we propose our method that resolves the above flaws.
To ensure that the transferred appearance aligns with the semantic meaning, we rearrange the features before the self-attention layer of the reference denoising process and inject it into the target process, which is intended to rearrange $Q_{\text{ref}}$.
Rearranging the reference features before the self-attention layer through precise correspondence matching and injecting them into the output denoising process yields better semantic alignment than KV injection.

\subsection{Semantic matching-based feature rearrangement}
\label{sec:matching}
\input{sources/math}
\input{sources/comparison}

As described in \secref{sec:motive}, previous appearance transfer methods with $KV$ injection do not always reflect \textit{semantic} correspondences between the reference and the target objects.
On the other hand, we explicitly rearrange the reference feature map to match the spatial arrangement of semantics with the target feature map. 

To find the semantic correspondence of a pixel $\mathbf{q}$ among the reference image at pixel $\mathbf{p}$, we take the $\argmax$ of the cosine similarity in the feature map before $l$-th self-attention layer at denoising timestep $t$. Additionally, to preserve the background of the target image, we apply object masks $M^{\text{target}}$ and $M^{\text{ref}}$ to the target and reference features:
\begin{equation} \label{eqn:mask}
F^\text{output}, F^\text{ref} = F^\text{output}[M^{\text{target}}], F^\text{ref}[M^{\text{ref}}],
\end{equation}
\begin{equation} \label{eqn:cossim}
\quad
\mathbf{p} = \argmax_{ \mathbf{p} \in [0, h) \times [0, w) }
                \text{sim}
                \left(
                    {F}^\text{output}(\mathbf{q}), {F}^\text{ref}(\mathbf{p})
                \right),
\quad
\end{equation}
where ${F}^\text{out} \in R^{hw \times c}$ and ${F}^\text{ref} \in R^{hw \times c}$ are the feature maps of the target and reference; $\text{sim}$ computes cosine similarity. $l$ and $t$ are omitted from ${F_{l,t}^*}$ for brevity. $\bigl[\cdot\bigr]$ denotes slicing by the object mask.

Then, we rearrange the reference feature map to reflect semantic correspondence, defining it as \(\Tilde{F}^\text{ref}(\mathbf{q}) = {F}^\text{ref}(\mathbf{p})\).  
This rearrangement is equivalent to modifying \(Q\) in self-attention.
Since $Q$ represents the object's structure, the rearrangement aligns the structure of the reference object to match the target object based on semantic matching.

Finally, we inject the rearranged reference features into the output denoising process. 
Reference features are aligned to the structure of the target object, enabling an effective transfer of the reference object's appearance.

\begin{align} \label{eqn:final}
{F'}^\text{output} = \tilde{F}^\text{ref}_{m} \odot M^{\text{target}}+F^\text{output} \odot (1-M^{\text{target}}),
\end{align}
where $\odot$ represents the Element-wise product.
\fref{fig:math} provides these processes of feature rearrangement and injection.
Then the self-attention becomes
\begin{align}
\text{softmax}\left(\frac{{Q'}^\text{output} ({{K'}^\text{output}})^T}{\sqrt{d}}\right) {V'}^\text{output},
\end{align}
where
${Q'}^\text{output}={F'}^\text{output}W^\text{query}$, ${K'}^\text{output}={F'}^\text{output}W^\text{key}$, and ${V'}^\text{output}={F'}^\text{output}W^\text{value}$.

Additionally, the transferred output often has different color brightness and contrast when compared to the reference.
To address this issue, we apply AdaIN\cite{huang2017arbitrary} used in Cross-Image \cite{cross_image} to masked noise, thereby reducing the color discrepancy between the reference and the output.

\input{sources/motivation2}
We next compare the matching processes of conventional methods and our method to highlight their differences.
Conventional matching methods \cite{zhang2024tale, dift, telling, luo2024diffusion} and ours find semantic correspondence by computing cosine similarity between two sets of feature pairs.
As illustrated in \fref{fig:comparison} (b), ours matches the reference features with \textit{the recursively transferred output features} and produces \textit{multiple matches} at multiple individual time steps.
It enables sparse key-point matching in the early step and dense matching in the later step, meaning that it can effectively capture both key-point and dense correspondence.
As shown in \fref{fig:motivation2} (c), while the early step flow map demonstrating dense matching is noisy compared to the ground truth (a), the sparse key point correspondences are accurate, and the flow map in the later step closely resembles the ground truth.
This later step's noise-free dense matching leads to a clean transferred result.
In contrast, conventional matching methods find matches between two fixed sets from reference and target features (\fref{fig:comparison} (a)).
Each fixed set forms \textit{a single set of matches}, either by aggregating features from multiple time steps \cite{luo2024diffusion} or by using features from an early time step \cite{dift, zhang2024tale}.
Semantic correspondence found with a single set of matches is suitable for finding sparse key-point matching at the RGB level, but inadequate for finding dense matching. 
In \fref{fig:motivation2} (b), the sparse key point correspondence in the early step is accurate, whereas the dense correspondence contains a lot of noise compared to the ground truth.
This noisy dense correspondence leads to a noisy transferred result.
With our improved dense matching, our transferred results are more realistic and have fewer artifacts than the ones from the transferred results using conventional methods, i.e., SD-DINO \cite{zhang2024tale}, DIFT \cite{dift} and TLFR \cite{telling}.

%% file: sources/motivation.tex
\definecolor{p_red}{RGB}{231, 50, 36}
\definecolor{p_green}{RGB}{56, 127, 35}
\definecolor{p_blue}{RGB}{23, 3, 242}
\definecolor{p_yellow}{RGB}{254, 251, 84}

\begin{figure}[t] \centering
    \includegraphics[width=\linewidth]{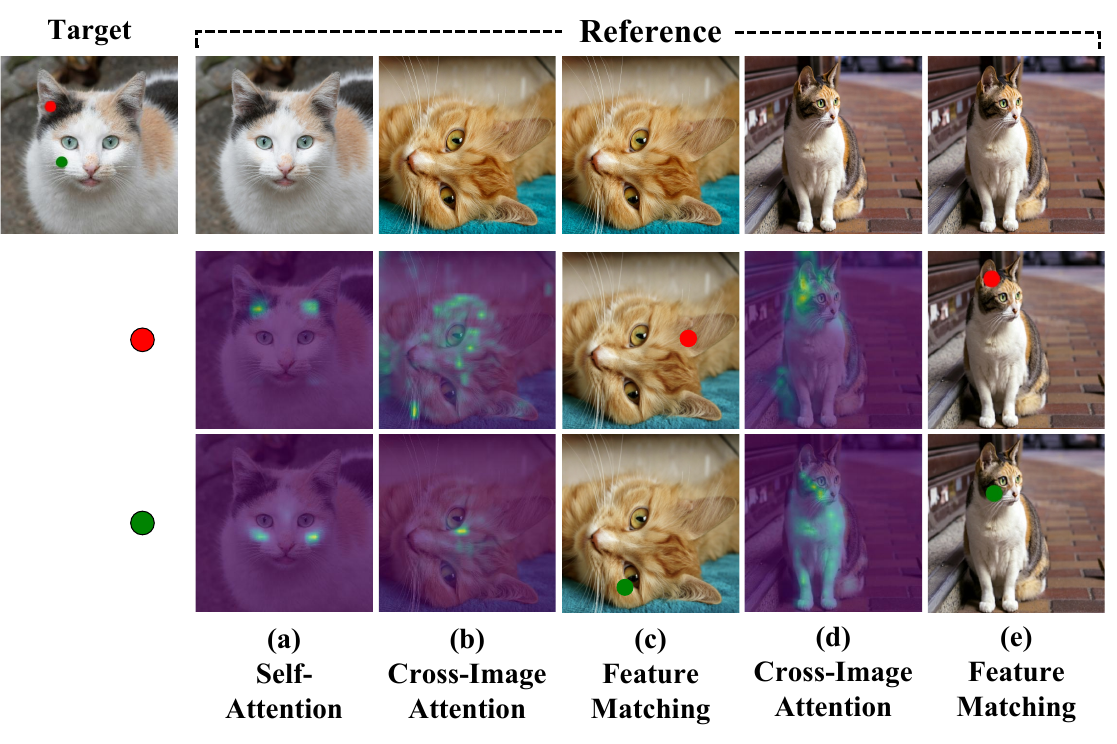}
    \caption{
        \textbf{Query-key attention maps vs. our feature matching.}
        For each query pixel $\mathbf{q}$ denoted by colored markers in the target image, we show the attention maps based on the $QK$ attention score. (b) and (d) include other regions in the attention map where matching is incorrect. In contrast, the feature matching in (c) and (e) presents a single point with the correct semantic meaning.}
    \label{fig:motivation}
    \vspace{-3mm}
\end{figure}

%% file: sources/math.tex
\begin{figure}[t]\centering
    \includegraphics[width=0.90\linewidth]{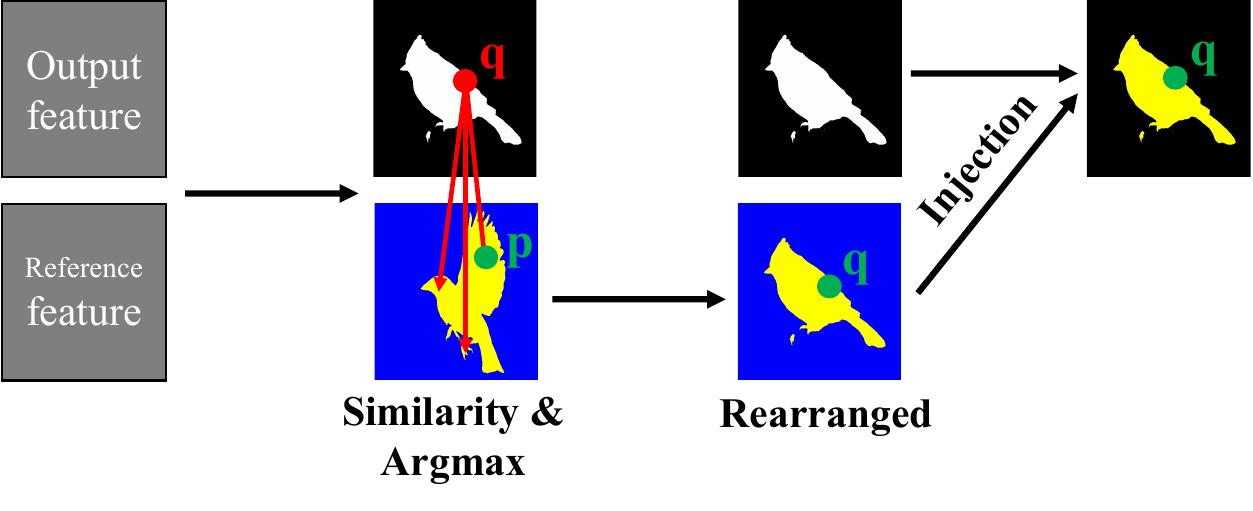}
    \vspace{-3mm}
    \caption{\textbf{Feature rearrangement and injection.} The reference feature, rearranged based on similarity to the output feature, is injected into the output denoising process.}
    \label{fig:math}
\vspace{-2mm}
\end{figure}

%% file: sources/comparison.tex
\begin{figure}[t]\centering
    \includegraphics[width=\linewidth]{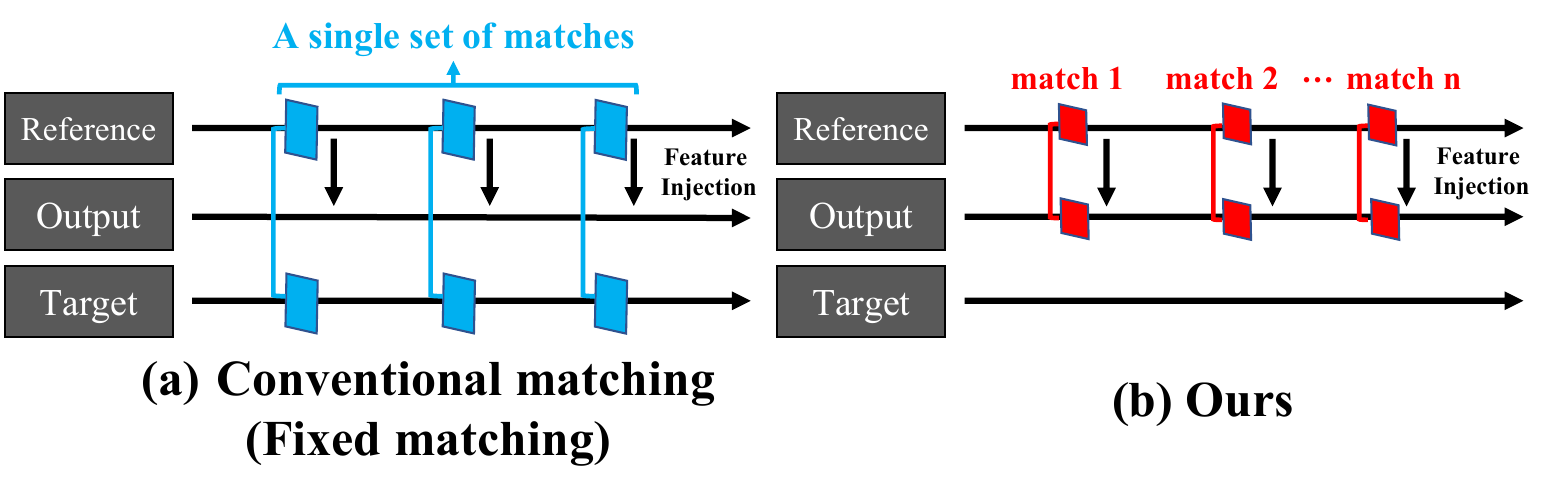}
    \vspace{-5mm}
    \caption{\textbf{Comparison between conventional matching methods and ours.} (a) Conventional methods aggregate features from multiple time steps of the reference and target into a single set and perform matching only once. (b) Ours matches the reference features with the output features and performs multiple matches across individual steps.}
    \label{fig:comparison}
\vspace{-3mm}
\end{figure}

%% file: sources/motivation2.tex
\begin{figure}[t] \centering
    \includegraphics[width=\linewidth]{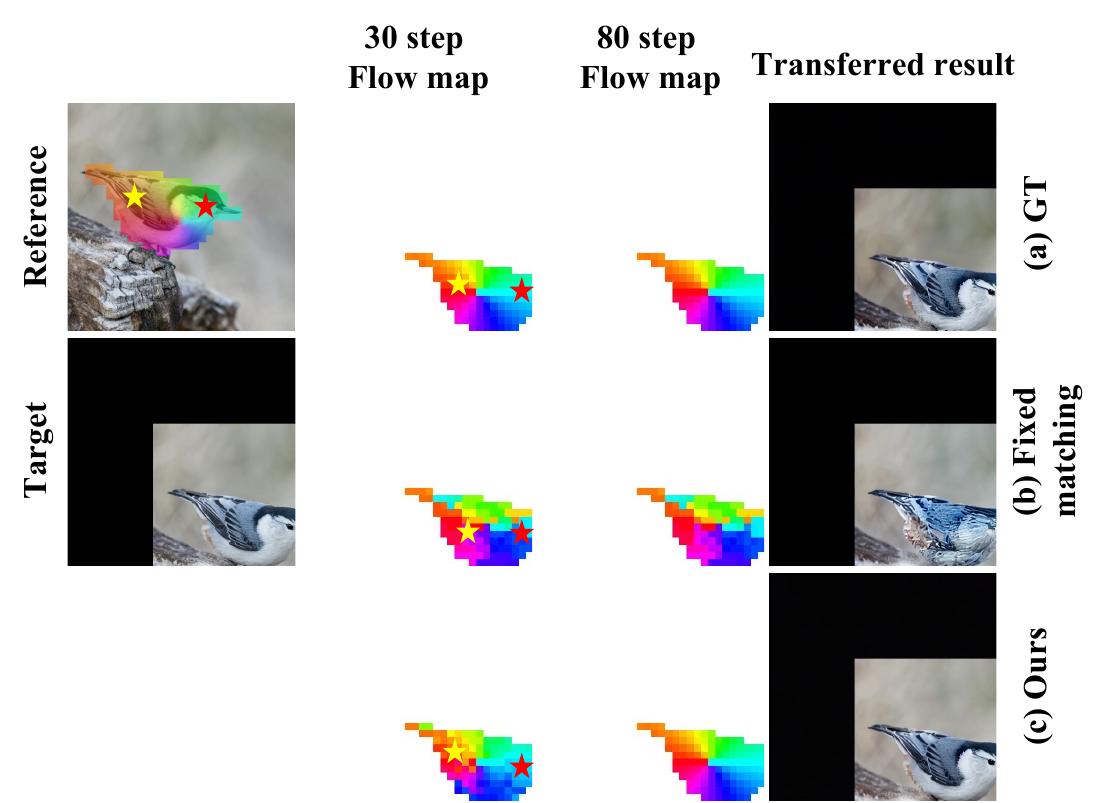}
    \caption{
        \textbf{Fixed matching vs. Our matching.}
        (a) Feature injection with Ground Truth (GT). GT represents the feature transferred from the reference. (b) Feature rearrangement and injection with fixed matching from conventional matching method \cite{dift}. (c) Feature rearrangement and injection at each step (ours). Our method performs dense matching closely aligned with the ground truth in later steps. Key-point matching at early time steps is represented as star markers.}
        \vspace{-3mm}
    \label{fig:motivation2}
\end{figure}

%% file: sec/4_experiments.tex
\input{sources/struct_prev}
\section{Experiments}

\myparagraph{Competitors} \label{exp:setting}
We compare our results with recent training-free diffusion-based methods, including Cross-Image \cite{cross_image}, which uses $KV$ injection, DiffEditor \cite{diffeditor}, which shows the best result among methods with score-based editing guidance \cite{self_guidance, dragon, diffeditor}, and DiffuseIT \cite{diffuseit}, which leverages external models for guidance.
In addition, we compare our method with the optimization-based approach Splice ViT \cite{spliceVIT}, the domain-specific trained Swapping Auto-Encoder (Swapping AE) \cite{swap_ae}, and also with IP-Adapter \cite{ip_adapter} and ZeST \cite{zest}, which adjust the appearance of the reference image based on a depth map input through Controlnet \cite{controlnet}.
In \aref{app:baseline}, we provide more details on implementation and hyperparameters for each method. 

\myparagraph{Evaluation Metric} \label{exp:setting}
In appearance transfer, the key evaluation factors are: 1) whether the appearance information from the reference is transferred to the correct location, and 2) whether the structure of the target object is well-preserved. For appearance evaluation, we assess the preservation of the overall color distribution by comparing the color histograms ($A_\text{hist}$) and evaluatethe semantic consistency by comparing the CLIP embeddings ($A_\text{clip}$) between the reference and transferred objects using the object masks. To evaluate structure preservation, we assess how much the structure in the generated result deviates from the target image using the following metrics. First, we evaluate semantic consistency by comparing key points ($S_\text{key}$) detected by ViT-Pose~\cite{vitpose} in the result image with the ground truth key points in the target. Also, to assess depth accuracy and object shape consistency, we calculate the RMSE of the depth maps ($S_\text{depth}$) and the mean intersection over union ($S_\text{miou}$) of the object masks~\cite{sam}. Finally, we measure dense correspondence using the method from a previous study \cite{zhang2024tale}, calculating the L1 distance of the flow map ($D_\text{flow}$). 
Additional details on evaluation metrics are provided in \aref{app:eval}, and explanations of the datasets used for evaluation can be found in \aref{app:dataset}.

\subsection{Appearance similarity}
\input{sources/result_4_2_1}
\input{sources/quan_table_app}
As shown in \fref{fig:qual}, our method successfully transfers the correct appearance from the reference to the target, even when the target and reference images are not aligned. 
For instance, in the bird examples, our results capture and reflect the complex patterns of the reference image, preserving the color arrangement of the blue head, green wings, and red-and-green body, while competitors fail to retain this arrangement. 
Furthermore, due to our method of rearranging features according to semantic meaning, the car and cat examples demonstrate our method’s robustness in cases where the reference and target are either unaligned or differ in size.
Also, our results accurately reflect the reference object's color, while IP-Adapter \cite{ip_adapter} and ZeST \cite{zest} generate unrealistic colors.
Notably, as shown in \tref{tab:quan_table_app}, ours achieves the lowest $A_\text{hist}$ score and the highest $A_\text{clip}$, highlighting its superior performance in preserving complex appearance patterns.

Additionally, as shown in  \fref{fig:various_domain}, our method successfully transfers the reference’s appearance across diverse domains, despite substantial differences domain between the reference and target images. 
Even when the reference and target belong to different domains, it successfully transfers the appearance of similar semantic meanings, such as a bird’s wing to an airplane’s wing.

\subsection{Structure preservation}
\input{sources/VvsF}
\input{sources/quan_table}

As shown in \tref{tab:quan_table}, our method achieves high performance in \( S_\text{depth} \), \( S_\text{miou} \), and \( S_\text{key} \), which evaluate structure preservation. 
Ours excel in both complex domains (e.g., buildings) and simpler ones (e.g., AFHQ), where the reference and target have similar sizes and poses.  
While IP-Adapter~\cite{ip_adapter} and ZeST~\cite{zest} may appear sufficient, their high scores result from using depth as an additional condition.
However, relying on such additional conditions to interpret the target structure can lead to incorrect estimations, especially when objects overlap or exhibit complex spatial arrangements, degrading object appearance transfer.
In contrast, our method achieves competitive or superior results without requiring extra depth information.
These quantitative results validate the structure-preserving capabilities observed in the qualitative examples in \fref{fig:qual}.
Cross-Image \cite{cross_image}, DiffEdit \cite{diffedit}, Splice VIT \cite{spliceVIT}, and IP-Adapter \cite{ip_adapter} produce results with altered target structures. 
The results highlight that ours notably outperforms in structure preservation and semantic consistency.

In appearance transfer, both appearance similarity and structure preservation are crucial. 
As shown in \fref{fig:qual}, \tref{tab:quan_table_app} and \tref{tab:quan_table}, compared to competitors that excel in only one aspect, ours achieves strong performance in both. 
It is possible due to the design of our method, which semantically rearranges the reference features to correspond with the target structure and injects them into object regions.

\subsection{Dense correspondence evaluation} 
\input{sources/dense_fig}
\input{sources/quan_table_dense}

In this section, we demonstrate, through dense correspondence evaluation, why it is essential to perform matching at each generation step in our method. 

\fref{fig:dense_figure} visualizes the correspondence between the reference feature map and the transferred feature map as a flow map.
As a baseline, we adopt Fixed Matching, where the matching rule is determined by a single set of matches and applied across all generation steps.
Our flow maps are smooth and free from noise, accurately reflecting the tendency of spatially adjacent pixels in an image to exhibit similar correspondences. 
In contrast, the Fixed Matching's flow maps contain more noise and lack smoothness. 
As a result, the transferred outputs from Fixed Matching exhibit significant noise and unnatural mismatched regions.

For quantitative evaluation, \tref{tab:quan_table_dense} compares the conventional semantic matching method with ours.
This is done using the optical flow smoothness metric \( D_\text{flow} \), as employed in the dense matching evaluation protocol~\cite{zhang2024tale}. 
Our method achieves a significantly lower flow map distance compared to other semantic correspondence methods.
Since the flow map distance increases when mismatching occurs or matching lacks smooth continuity, it indicates that our method demonstrates the highest dense correspondence performance among existing matching methods based on diffusion features. Please refer to \aref{app:eval} for the evaluation details.

\subsection{Analysis of rearrange and injection component} \label{sec:injection}

In this section, we demonstrate that our feature injection aligns with human intuition.
\fref{fig:attn_map} (a) shows an example of appearance transfer that aligns with human expectations by considering semantic meaning; for instance, transferring the appearance of the reference belly to the target belly. Furthermore, humans tend to interpret even within the same region by segmenting colors, as in (b), rather than allowing arbitrary matches within a semantic region. \fref{fig:attn_map} (c) demonstrates that our feature injection \texttt{Ours} process aligns with human intuition, unlike key-value injection (\texttt{on KV}) and value injection (\texttt{on V}).

\texttt{Ours} produces a clear, lime-like belly that accurately reflects the reference, while \texttt{on KV} and \texttt{on V} result in red smudges in the belly. We demonstrate this effect in \fref{fig:attn_map} (c) by visualizing attention maps across denoising timesteps, where each attention map corresponds to the activation map for the red dot in the target image. \texttt{Ours} aggregates visual elements from the lime-colored reference belly according to its semantics. In contrast, \texttt{on KV} and \texttt{on V} focus on the red neck and head, disregarding the belly’s semantics.

We suggest the reason as follows. 
Our method rearranges the reference $F$ according to the semantic correspondence to the target and replaces the target $F$ with it. 
Hence, our results have the visual elements of the reference arranged in the semantic structure of the target. 
In contrast, the $Q$ in \texttt{on KV} and \texttt{on V} assigns high attention to the red color on the reference belly (yellow boxes in \fref{fig:attn_map} (c)) and transfers it to the orange-ish belly of the target bird. 
It occurs because, unlike in \texttt{ours}, where the target $Q$ is replaced with the rearranged reference $Q$, \texttt{on KV} and \texttt{on V} retain the original target $Q$, causing the target bird's belly color to be interpreted as a different structure (\fref{fig:attn_map} (b)).

\subsection{Ablation study}
\input{sources/ablation}

In this section, we perform ablation experiments regarding different components of our method and show its contribution in \fref{fig:ablation}.
Compared to $KV$ injection (c), our semantic matching-based feature rearranging (d) transfers appearance to regions where the semantic meaning of objects aligns. 
For instance, unlike (c), where the reference car's headlights are transferred to the side of the target car, (d) correctly transfers the side of the reference car to the side of the target car, resulting in a properly transferred black car.
In (e), the AdaIN on masked noise matches the global color distribution of the object, thereby maintaining the color brightness and contrast of the appearance object. We provide quantitative ablation results and more various ablation studies in \aref{app:feature_position} to  \aref{app:ablation_matching_rule}.

\subsection{Application}
\input{sources/application}
\myparagraph{Cross-style appearance transfer}
Our method enables semantic matching-based transfer even in challenging samples where the target object and the reference object belong to exhibit different styles. In \fref{fig:application}, (a) depicts the appearance of a real rabbit applied to a Disney-style rabbit.

\myparagraph{Multi-objects appearance transfer} Our method can individually transfer the appearance of multiple objects in the target image, each from a different reference image.
Objects in the target image are matched and rearranged one by one with the reference images.
Each process is executed simultaneously within a single generation process, rather than sequentially. In \fref{fig:application}, (b) presents the results with three birds in the target image, each with distinct appearances from three different images.

%% file: sources/struct_prev.tex
\begin{figure*}[t] \centering
    \addtolength{\belowcaptionskip}{-5pt}
    \includegraphics[width=\linewidth]{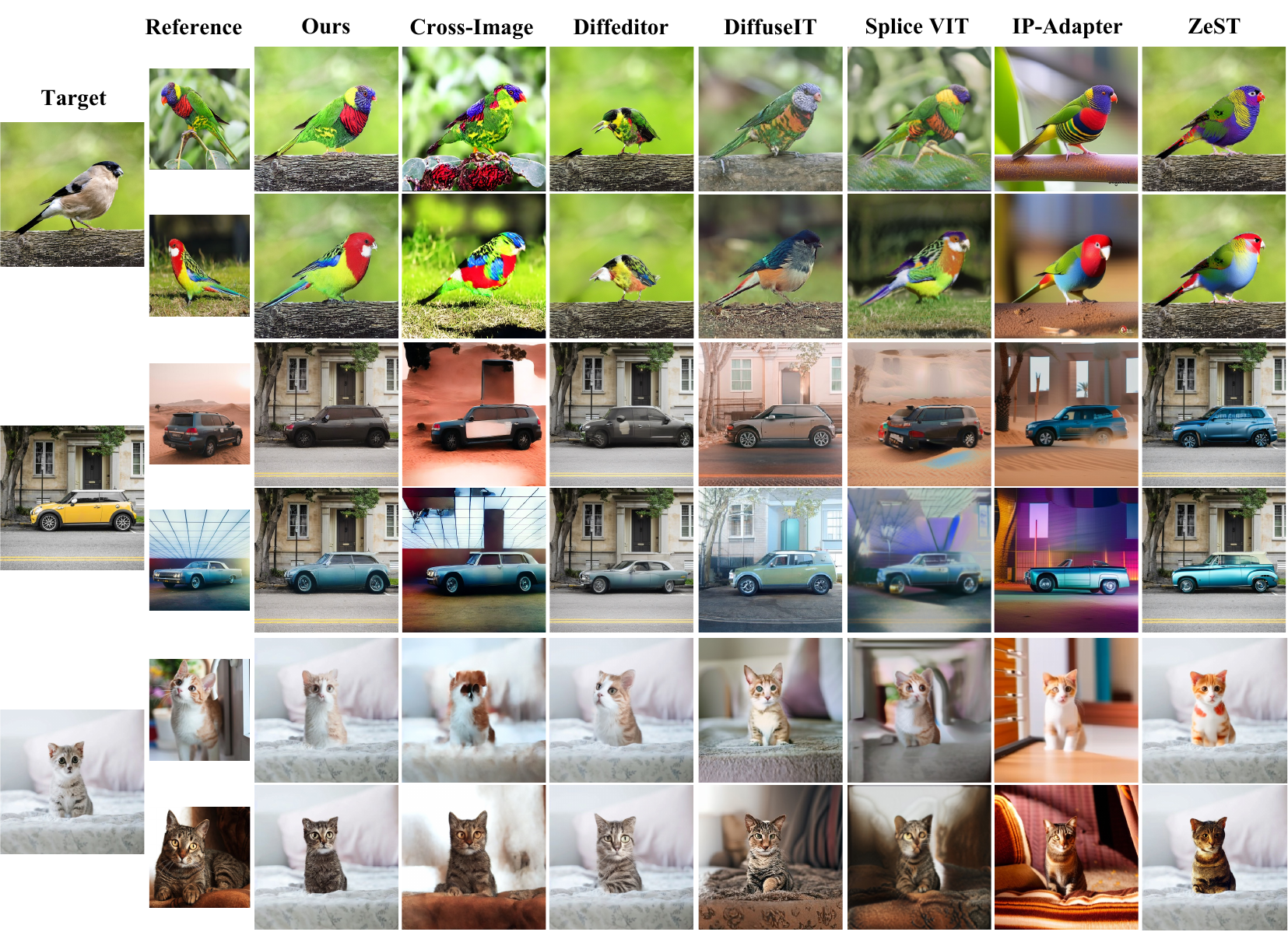}
    \caption{\textbf{Qualitative comparison.} We compare our results with the competitors on samples where the target and reference objects are unaligned, have complex patterns, or differ in size. The competitors struggle in various ways.}
    \label{fig:qual}
\end{figure*}

%% file: sources/result_4_2_1.tex
\begin{figure}[t] \centering
    \addtolength{\belowcaptionskip}{-5pt}
    \includegraphics[width=\linewidth]{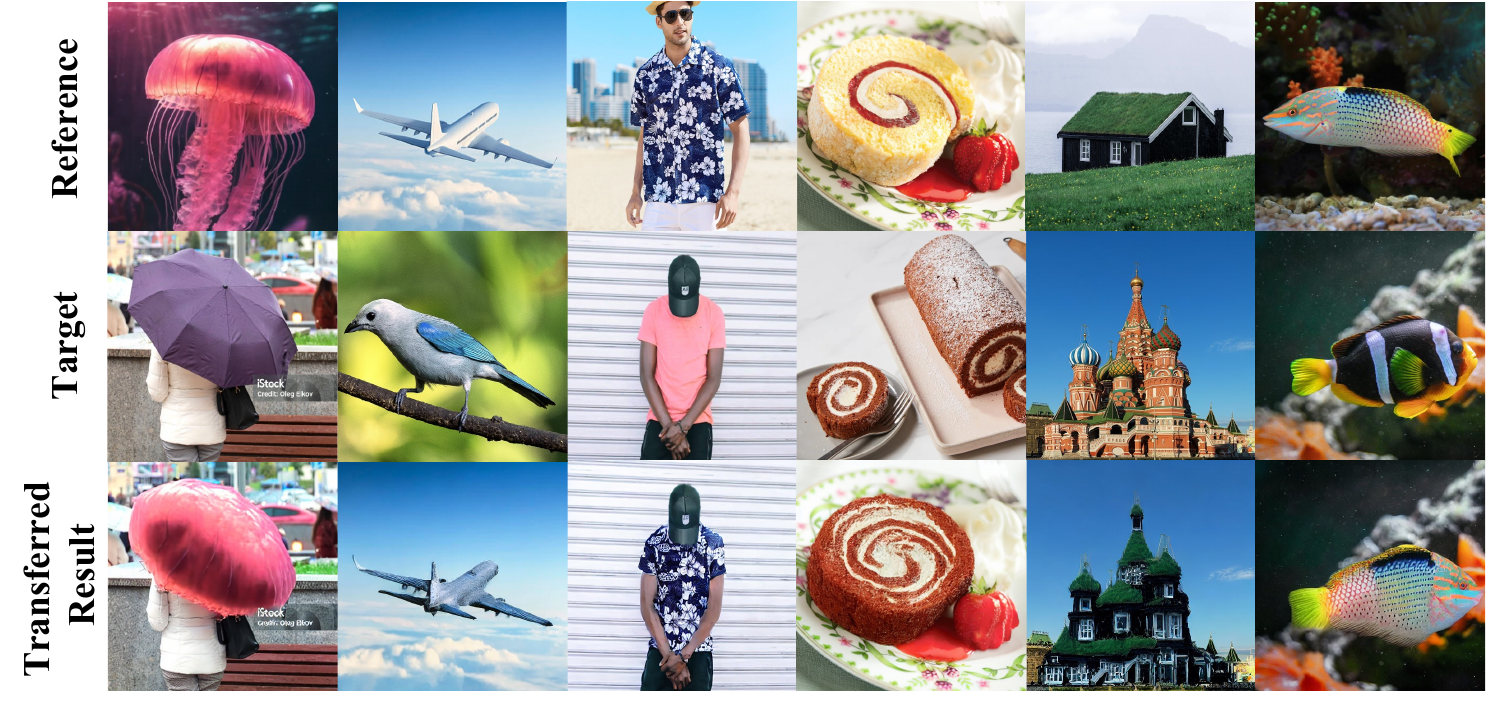}
    \caption{\textbf{Our results of various domain.} Our approach can transfer appearance across diverse domains. }
    \label{fig:various_domain}
    \vspace{-3mm}
\end{figure}

%% file: sources/quan_table_app.tex
\begin{table}[]
\centering
\resizebox{0.9\linewidth}{!}{
\tiny  
\begin{tabular}{c|cc|cc}
\specialrule{.05em}{1em}{0em} 
\hline
\multicolumn{1}{c|}{Method} &
\multicolumn{2}{c|}{${A_\text{hist}} \downarrow$} & 
\multicolumn{2}{c}{${A_\text{clip}} \uparrow$} \\
\hline 
\multicolumn{1}{c|}{Dataset} &
\multicolumn{1}{c}{Building} &
\multicolumn{1}{c|}{AFHQ} &
\multicolumn{1}{c}{Building} &
\multicolumn{1}{c}{AFHQ} \\
\hline
Ours             & \cellcolor[HTML]{FFCCC9}0.469 & \cellcolor[HTML]{FFCCC9}0.577 & \cellcolor[HTML]{FFCCC9}95.30 & \cellcolor[HTML]{FFCCC9}97.03 \\ 
Cross-Image      & 0.491  & 0.608 & 94.05 & 96.75 \\
DiffEditor       & 0.478  & 0.614 & 91.35 & 96.13 \\
DiffuseIT        & 0.477  & 0.607 & 90.74 & 96.21 \\
Splice ViT       & \cellcolor[HTML]{FFFFC7}0.472  & \cellcolor[HTML]{FFFFC7}0.580 & \cellcolor[HTML]{FFFFC7}94.64 & 96.30 \\
Swapping AE      & 0.481  & 0.629 & 85.90 & 92.35 \\ 
IP-Adapter       & 0.487  & 0.616 & 93.46 & \cellcolor[HTML]{FFFFC7}96.76 \\ 
ZeST             & 0.497  & 0.602 & 89.59 & 96.04 \\ 
\hline
\end{tabular}
}
\caption{\textbf{Quantitative evaluation for appearance similarity.} 
We mark the best score in red and the second-best score in yellow.}
\vspace{-3mm}
\label{tab:quan_table_app}
\end{table}

%% file: sources/VvsF.tex
\begin{figure*}[t] \centering
    \includegraphics[width=0.9\linewidth]{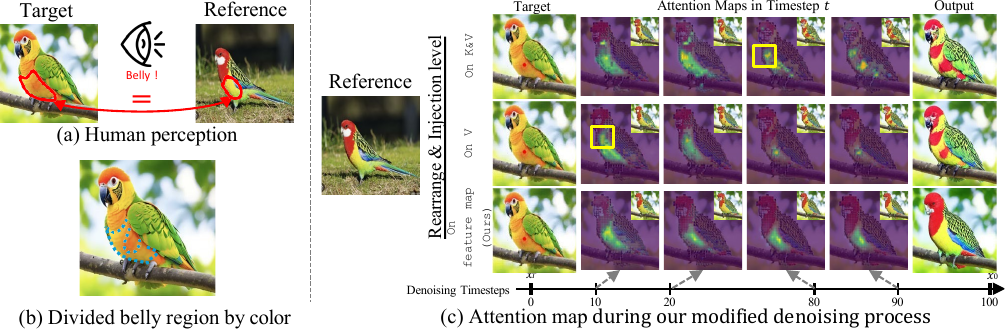}
    \caption{\textbf{Comparison of attention maps.} 
    (a) shows the corresponding region between the target and the reference from human perception. (b) illustrates dividing regions based on color, not semantic meaning. (c) provides the attention maps for the target image's query pixel (red dot) at different timesteps during appearance transfer. The K\&V modification (first row) and V modification (second row) perform semantic matching in the same manner as our method but apply the rearrangement and injection processes to K\&V and V instead of the feature map, respectively. The image at the top right of each attention map represents the result of feature rearrangement, which is indirectly shown by rearranging the reference RGB image with semantic matching calculated from U-Net's 2nd up-block.
    }
    \label{fig:attn_map}
    \vspace{-1.5mm}
\end{figure*}

%% file: sources/quan_table.tex
\begin{table}[]
\centering
\resizebox{\linewidth}{!}{
\scriptsize  
\begin{tabular}{c|cc|cc|c}
\specialrule{.05em}{1em}{0em} 
\hline
\multicolumn{1}{c|}{Method} & \multicolumn{2}{c|}{${S_\text{depth}} \downarrow$} & \multicolumn{2}{c|}{${S_\text{miou}} \uparrow$} &
\multicolumn{1}{c|}{${S_\text{key}} \uparrow$}\\ \hline
\multicolumn{1}{c|}{Dataset} & \multicolumn{1}{c}{Building} & \multicolumn{1}{c|}{AFHQ} & \multicolumn{1}{c}{Building} & \multicolumn{1}{c|}{AFHQ} & \multicolumn{1}{c}{AP-10K} \\ \hline
Ours             & \cellcolor[HTML]{FFCCC9} 0.197 & \cellcolor[HTML]{FFCCC9} 0.114 &
                \cellcolor[HTML]{FFCCC9} 0.939 & \cellcolor[HTML]{FFFFC7} 0.972 & 
                \cellcolor[HTML]{FFFFC7} 82.99 \\ 
Cross-Image      & 0.287  & 0.139 & 0.758 & 0.915 & 64.49 \\
DiffEditor       & 0.266  & 0.124 & 0.863 & 0.943 & 46.26 \\
DiffuseIT        & 0.263  & 0.123 & 0.855 & 0.951 & 65.07 \\
Splice ViT       & 0.319  & 0.120 & 0.842 & 0.943 & 47.54 \\
Swapping AE      & 0.295  & 0.128 & 0.821 & 0.942 &  N/A \\ 
IP-Adapter       & 0.374  & 0.130 & 0.642 & 0.950 & \cellcolor[HTML]{FFCCC9} 84.26 \\ 
ZeST             & \cellcolor[HTML]{FFFFC7} 0.242 & \cellcolor[HTML]{FFFFC7} 0.119 &
                \cellcolor[HTML]{FFFFC7} 0.925 & \cellcolor[HTML]{FFCCC9} 0.980 & 78.25 \\ 
\hline
\end{tabular}
}
\caption{\textbf{Quantitative evaluation for structure preservation.} We mark the best score in red and the second-best score in yellow.}
\label{tab:quan_table}
\vspace{-3mm}
\end{table}

%% file: sources/dense_fig.tex
\begin{figure}[t] \centering
    \includegraphics[width=\linewidth]{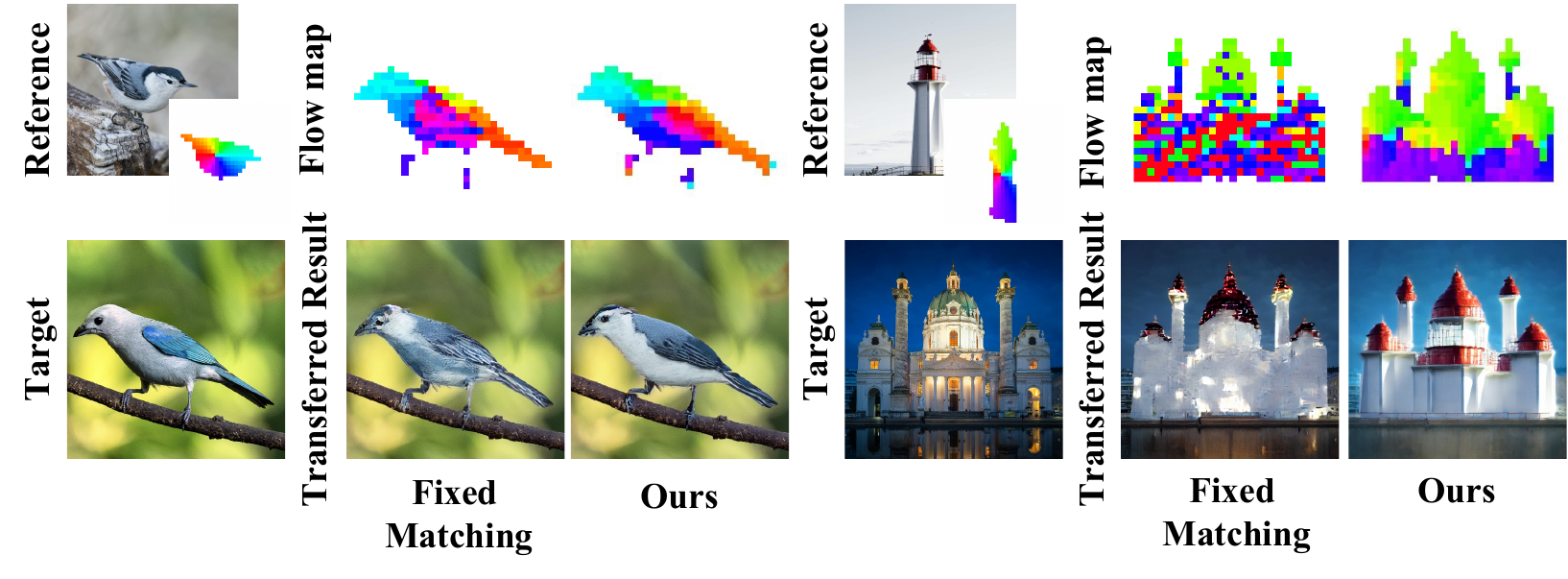}
    \vspace{-5mm}
    \caption{
        \textbf{Qualitative comparison between the result of fixed matching and ours.} The first row displays the reference color map and the flow map. The flow map shows the result of mapping the colors of reference pixels to  matched target pixels. The second row compares the transfer result using fixed matching from an existing semantic matching method \cite{zhang2024tale} with our transferred result.}
    \label{fig:dense_figure}
    \vspace{-3mm}
\end{figure}

%% file: sources/quan_table_dense.tex
\begin{table}[t]
\centering
\resizebox{0.85\linewidth}{!}{
\scriptsize  
\begin{tabular}{c|ccc}
\specialrule{.05em}{1em}{0em} 
\hline
\multicolumn{1}{c|}{Method} & \multicolumn{3}{c}{${D_\text{flow}} \downarrow$} \\ \hline
\multicolumn{1}{c|}{Dataset} & \multicolumn{1}{c}{FG3D CAR
} & \multicolumn{1}{c}{JODS} & \multicolumn{1}{c}{PASCAL} \\ \hline
Ours             & \cellcolor[HTML]{FFCCC9}9.43 & \cellcolor[HTML]{FFCCC9}28.75 & \cellcolor[HTML]{FFCCC9}21.83\\ 
TLFR\cite{telling} *  & 41.11 & 65.28 & 106.53 \\
TLFR\cite{telling}    & 30.75 & 59.85 & 102.14 \\
SD-DINO\cite{zhang2024tale} & \cellcolor[HTML]{FFFFC7}26.87 & \cellcolor[HTML]{FFFFC7}47.54 & \cellcolor[HTML]{FFFFC7}63.27 \\
DIFT\cite{dift} & 77.53 & 91.32 & 135.37 \\
\hline
\end{tabular}
}
\caption{\textbf{Quantitative evaluation for dense matching.} * indicates a fine-tuned backbone. We mark the best score in red and the second-best score in yellow.}
\label{tab:quan_table_dense}
\vspace{-3mm}
\end{table}

%% file: sources/ablation.tex
\begin{figure}[t]\centering
    \includegraphics[width=0.93\linewidth]{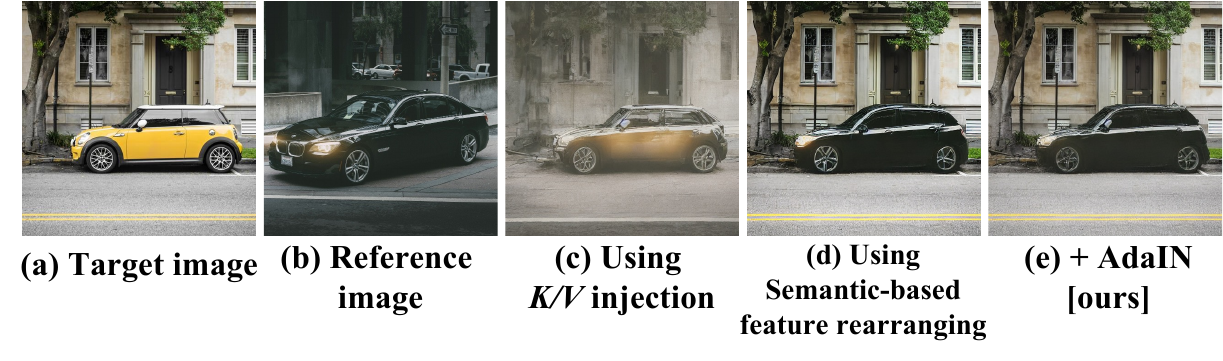}
    \vspace{-1.5mm}
    \caption{We perform an ablation study to validate our method.}
    \label{fig:ablation}
\vspace{-3mm}
\end{figure}

%% file: sources/application.tex
\begin{figure}[t]\centering
    \includegraphics[width=\linewidth]{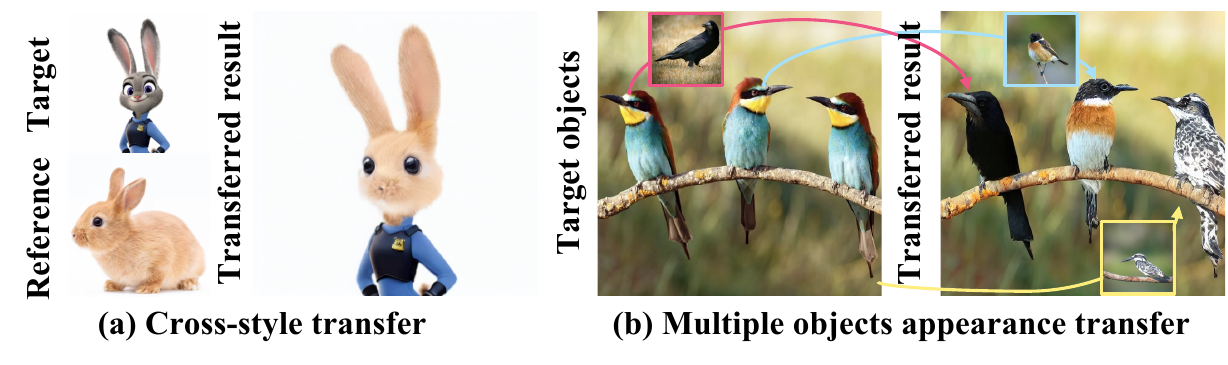}
    \vspace{-6mm}
    \caption{Results of various applications}
    \label{fig:application}
    \vspace{-1mm}
\end{figure}

%% file: sec/5_conclusion.tex
\section{Conclusion}

\begin{figure}[t] \centering
    \includegraphics[width=\linewidth]{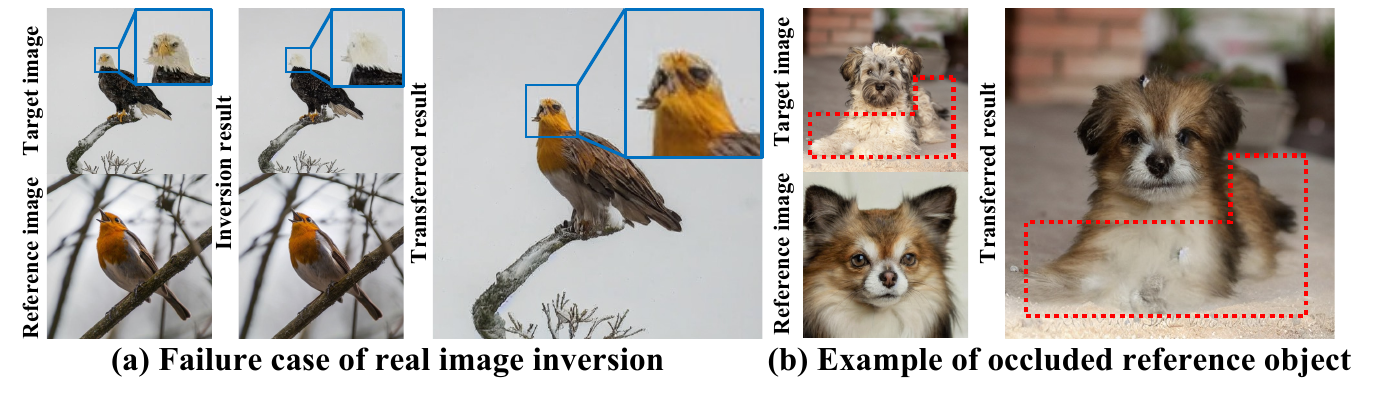}
    \vspace{-6mm}
    \caption{
        Results with limitations
    }
    \label{fig:limitation}
     \vspace{-3mm}
\end{figure}

In this paper, we have introduced a semantic-based appearance transfer method using a pretrained text-to-image diffusion model. Our method faithfully reflects the reference image to the target image according to semantic correspondences, e.g., fin-to-fin and wing-to-wing, while previous methods often ignore semantics. Our key arguments for replacing features in the target denoising process with the reference denoising process are 1) reflecting dense semantic correspondences 2) found during the modified denoising process 3) on the input features of self-attention. Experiments demonstrate that our method achieves faithful appearance transfer between the semantically corresponding parts of the result and the reference and better preserves the structure of the target in the result compared to existing methods. Furthermore, we achieve significantly superior dense semantic correspondence results compared to existing semantic matching methods.

\paragraph{Limitation}
In order to use a real image as a reference, our method relies on inversion. If the inversion malfunctions, our method struggles as shown in \fref{fig:limitation} (a).
Also, \fref{fig:limitation} (b) shows that the reference image does not have the semantically corresponding parts from the target image and our matching finds the most similar parts instead of semantic correspondence.
Still, the results tend to be realistic in such cases.

%% file: sec/6_appendix.tex
\begin{appendix}

\renewcommand{\thetable}{S\arabic{table}}
\renewcommand{\thefigure}{S\arabic{figure}}
\setcounter{figure}{0}
\setcounter{table}{0}

\section{Implementation details} \label{app:implementation}
We apply the proposed methods to the text-to-image diffusion model, Stable Diffusion \cite{stable} using checkpoint v1.5. 
We begin by inverting real images with the edit-friendly DDPM inversion \cite{edit_friendly}, sampling images with 100 denoising timesteps. 
To find semantic correspondence during transfer, we use the feature maps input to the self-attention layer. 
We set the denoising step t ∈ [42, 100] and layer l ∈ [2, 3] from the up-blocks of U-net to find correspondences and rearrange features. 
Additionally, we apply AdaIN at denoising step t ∈ [82, 100] and use the off-the-shelf model SAM \cite{sam} to obtain object masks. 
And we measure dense correspondence at timestep 92 and layer 2.
All of the experiments are conducted on an NVIDIA A6000 GPU and during the transfer experiments, the GPU memory usage amounted to about 15.17 GB.

\section{Evaluation method details} \label{app:eval}
\subsection{Appearance similarity}
To evaluate the success of transferring the appearance of the reference image, we conduct an experiment comparing the color histograms ($A_\text{hist}$) of the result image and the ground truth (GT) image. 
The comparison region is set by segmenting the object using SAM \cite{sam} for both the GT and result images.
For the comparison of color histograms, we measure the Bhattacharyya distance as:

\begin{equation}
    A_\text{hist}(H_G, H_O) = D_B(H_G, H_O)
\end{equation}

where \( D_B(H_G, H_O) \) is the Bhattacharyya distance between the color histograms of the masked GT image (\( H_G \)) and the masked transferred output image (\( H_O \)).

Additionally, we measure semantic similarity using CLIP score:

\begin{equation}
    A_{\text{clip}}(G, O) = \frac{1}{N} \sum_{i=1}^{N} \text{CLIP}(G_i, O_i)
\end{equation}

where  
\( G_i \) and \( O_i \) are the masked GT and masked transferred output images, respectively,  
and \( N \) is the total number of images.

The dataset used in the experiments is described in \aref{app:dataset}.

\subsection{Structure preservation}
To evaluate the preservation of the target image's structure, we conduct a depth evaluation (\( I_\text{depth} \)), a mean Intersection over Union (mIoU, \( S_\text{miou} \)) and a key point evaluation (\( S_\text{key} \)).

For \( S_\text{depth} \), we use an off-the-shelf depth estimation model~\cite{MiDas}.
We extract depth from the target image and the transferred results of each model, then measure the root mean square error (RMSE) at the object level:

\begin{equation}
    S_\text{depth}(D_T, D_O) = \sqrt{ \frac{1}{N} \sum_{i=1}^{N} \left( D_{T, i} - D_{O, i} \right)^2 }
\end{equation}

where  
\( D_T \) and \( D_O \) are the depth maps of the masked target image and the transferred output image, respectively,  
and \( N \) is the total number of pixels.

For \( S_\text{miou} \), we use SAM to obtain the masks of the ground truth (GT) and the transferred result objects.
The mIoU is then measured at the object level as:

\begin{equation}
    S_\text{miou}(T, O) = \frac{1}{N} \sum_{i=1}^{N} \frac{|M_{T, i} \cap M_{O, i}|}{|M_{T, i} \cup M_{O, i}|}
\end{equation}

where  
\( T \) and \( O \) denote the target and output images,  
\( M \) represents the object mask obtained from SAM-HQ,  
and \( N \) is the total number of objects.

To follow the default settings of the models, ours, Cross-Image \cite{cross_image}, DiffEditor \cite{diffeditor}, Splice VIT \cite{spliceVIT}, IP-Adapter \cite{ip_adapter}, and ZeST \cite{zest} are tested at an image resolution of \( 512^2 \).
Swapping AE \cite{swap_ae} and DiffuseIT \cite{diffuseit} are tested at a resolution of \( 256^2 \).

For \( S_\text{key} \), we assess structural preservation through pose estimation with ViTPose++ \cite{vitpose}.
Following its approach, we evaluate AP-10K samples [AP-10K] from the training set and compute Average Precision (AP) using Object Keypoint Similarity (OKS) over thresholds $\tau \in [0.5, 0.95]$ with target keypoints as ground truth.
Our method achieves higher AP than competitors, demonstrating superior structural retention. OKS is defined as:

\begin{equation}
    \text{OKS} = \frac{\sum_i \exp\left(-\frac{d_i^2}{2s^2\kappa_i^2}\right) \delta(v_i > 0)}{\sum_i \delta(v_i > 0)},
\end{equation}

where $d_i$ is the Euclidean distance between the detected and ground truth keypoints, $s$ is the object scale, $\kappa_i$ is a keypoint-specific constant, and $v_i$ is the keypoint visibility.

Using OKS, the Average Precision (AP) score is computed as:

\begin{equation}
    \text{AP} = \frac{1}{|\tau|} \sum_{\tau} \text{Precision}(\tau).
\end{equation}

The precision at each threshold \( \tau \) is given by:

\begin{equation}
    \text{Precision}(\tau) = \frac{|\{ \text{detected keypoints} \mid \text{OKS} \geq \tau \}|}{|\{ \text{all detected keypoints} \}|}.
\end{equation}

The dataset used in the experiments is described in \aref{app:dataset}.

\subsection{Dense correspondence}
We evaluate dense correspondence using flow maps, which represent pixel displacements derived from the correspondences estimated by each method.
These flow maps are computed by subtracting the difference between the target pixel coordinates from their corresponding matches.
To measure deviations from the GT flow map, we calculate the L1 distance at the resolution of $512^2$ as, 

\begin{equation}
    D_{\text{flow}}(F_{\text{pred}}, F_{\text{GT}}) = \frac{1}{N} \sum_{i=1}^{N} \frac{\sum |F_{\text{pred}, i} - F_{\text{GT}, i}|}{|\mathcal{M}_i|}
\end{equation}

where  
\( N \) is the total number of images,  
\( F_{\text{pred}, i} \) and \( F_{\text{GT}, i} \) are the predicted and ground truth optical flow for image \( i \), respectively,  
and \( \mathcal{M}_i \) is the validity mask indicating the valid pixels in the flow.

SD-DINO \cite{zhang2024tale} and Telling-Left-from-Right \cite{telling} employ both $960^2$ and $840^2$ image resolutions to extract feature descriptors across two distinct models, and ours utilizes a resolution of $512^2$.
Source and target images of varying sizes are resized to the input resolution required by each method, following the padding strategy detailed in the official implementation of SD-DINO \cite{zhang2024tale}.
Both ours and SD-DINO \cite{zhang2024tale} compute dense correspondence by upsampling feature maps to $512^2$.
Telling-Left-from-Right \cite{telling} derives dense correspondence with feature maps at their original resolution ($60^2$), using a window-soft-argmax operation, and subsequently upsamples the correspondence map to $512^2$.
The dataset used in the experiments is described in \aref{app:dataset}.

\section{Evaluation dataset} \label{app:dataset}
\input{sources/suppl/augment_data}

For the quantitative evaluation, we used the AFHQ \cite{Starganv2}, AP-10K \cite{ap10k}, and TSS \cite{tss} datasets, and a Building dataset collected from the Pexels\footnote{https://www.pexels.com/}.
This dataset will be publicly available.
Especially, as shown in \fref{fig:augment_data}, to evaluate appearance transfer performance, we created datasets for $A_\text{hist}$ and $A_\text{clip}$ with the following setup: (1) Reference: original image (2) Target: shape and color-augmented image derived from the original image (3) Ground-Truth (GT): shape augmented image derived from the original image. We perform appearance transfer on 1) Reference to (2) Target, and measure the score by comparing the result object with (3) GT object. To align with the training domain of the pre-trained Swapping AE \cite{swap_ae}, we applied flip and weak warping as augmentations. 
Additionally, to evaluate structure preservation, we use a building dataset comprising 30 pairs of structure and target images, as well as an AFHQ dataset with 42 pairs.
We evaluate dense correspondence on the TSS dataset \cite{tss}, which includes dense correspondence flows and semantic masks for 400 image pairs sampled from the FG3DCAR \cite{fg3dcar}, JODS \cite{jods}, and PASCAL \cite{pascal} datasets.

\section{Baseline settings} \label{app:baseline}

All experiments were conducted at a resolution of $512^2$, except for the Swapping AE and DiffuseIT, which were trained at a resolution of $256^2$.

\subsection{For appearance transfer comparison}
\paragraph{Cross-Image.} 
Cross-Image \cite{cross_image} employs edit-friendly DDPM inversion \cite{edit_friendly} for image inversion. 
Images are sampled with 100 denoising timesteps. 
And Cross-Image does not use an object mask during transfer, so the background of the target is not preserved after the transfer.
The KV injection in self-attention occurs at t ∈ [42, 100] and layer l ∈ [2, 3] from the up-blocks. 
The contrast strength is set to 1.65, and the swap guidance scale is set to 3.5. 
Additionally, for consistency with our model, experiments were conducted using Stable Diffusion v1.5. 
\paragraph{DiffEditor.} 
DiffEditor is experimented with under Stable Diffusion v1.5. 
We use the standard DDIM scheduler for 50 denoising steps. 
The classifier-free guidance scale was set to the default value of 5.
And Diffeditor uses an object mask during transfer, so the background of the target is preserved.
\paragraph{DiffuseIT.} 
DiffuseIT \cite{diffuseit} utilizes external models to guide the denoising process. 
We set the denoising timestep to 200, skipping the initial 80 timesteps, and use a resampling step of N=10 (resulting in a total of 130 iterations). 
Images are resized to a resolution of $224^2$ to compute the ViT and CLIP losses, as these models only accept this resolution. 
These settings are the default configuration for image-guided manipulation as specified by the authors. 
Additionally, other configurations, including hyperparameters, follow the default settings provided by the authors. 
Since the provided checkpoint is trained at a resolution of $256^2$, we also conducted experiments at this resolution.
\paragraph{Splice ViT.} Splice ViT \cite{spliceVIT} employs a pre-trained DINO ViT model \cite{vitpose} as a feature extractor for optimizing the model on a single image pair. 
We use the 12-layer pre-trained ViT-B/8 model provided in the official DINO ViT implementation. 
For the ViT loss, images are resized to a resolution of $224^2$. 
Keys are derived from the deepest attention module for self-similarity, and the output of the deepest layer is used to extract the appearance from the target appearance image. 
We optimize using an input image pair with a resolution of $512^2$ for 2000 iterations. 
These settings follow the default settings provided by the authors, and other configurations, including hyperparameters, also follow the provided configurations.
\paragraph{Swapping AE.} We use the pretrained checkpoints provided on the official GitHub repository. 
We evaluate the AFHQ dataset and the LSUN Church pretrained models, treating the Building dataset as in-domain for LSUN Church model. 
In all evaluations, the target image is treated as the structure image, and the reference image is treated as the texture image. 
Additionally, we set the texture mixing alpha to 1.0, i.e,. simple texture swapping.
\paragraph{IP-Adapter.}
To account for target depth, we adopt the IP-Adapter + ControlNet model, using SDXL as the base model. The target image’s depth map is extracted using off-the-shelf depth estimator~\cite{MiDas}, normalized, and then used as a condition for ControlNet. The reference image is provided as the input image. The ControlNet conditioning scale is set to 0.7, and the DDIM step is set to 30, following the inference settings from the official repository.
\paragraph{ZeST.}
ZeST utilizes Dense Prediction Transformers \cite{DPT} for depth estimation and Rembg \cite{rembg} for foreground extraction. 
It also employs Stable Diffusion XL Inpainting in conjunction with the corresponding version of depth-based ControlNet and IP-Adapter. 
Additionally, all other configurations, including hyperparameters, follow the default settings provided by the authors.

\subsection{For semantic correspondence comparison}
\paragraph{SD-DINO.} SD-DINO \cite{zhang2024tale} employs Stable Diffusion v1.5 with a diffusion model timestep of $t=100$ as the visual descriptor, while integrating DINOv2 \cite{dino} as an auxiliary descriptor.
Stable Diffusion features are extracted from the 2nd, 5th, and 8th layers of the U-Net decoder at timestep $t=50$, while DINOv2 descriptors are derived from the token facet of its 11th layer.
The input resolutions are $960^2$ for Stable Diffusion and $840^2$ for DINOv2, resulting in a feature map with a resolution of $60^2$.
Then, we use $512^2$ upsampled feature map to find semantic correspondence.
\paragraph{Telling-Left-from-Right.} Telling-Left-from-Right \cite{telling} adopts Stable Diffusion and DINOv2 features in a manner similar with SD-DINO.
Furthermore, it incorporates the instance matching distance (IMD) to compare the target image with the horizontally flipped source image, thereby mitigating pose variation in paired images.
Semantic correspondence is computed on the $60^2$ resolution map using window soft argmax with a window size of 10, followed by upsampling to $512^2$ for evaluation.

\section{Ablation study for feature positions}
\label{app:feature_position}
\input{sources/suppl/more_layer}
\input{sources/suppl/abl_layer}
We use the input features of the self-attention layer for correspondence measurement and feature injection. 
However, the output of the self-attention layer can also be used for correspondence measurement. 
Through experiments, we confirm that the input features to self-attention yield better performance. 
\fref{fig:more_layer} (a), which uses the self-attention output features, shows less accurate matching compared to \fref{fig:more_layer} (b), which uses the self-attention input features.
And as shown in \tref{tab:quan_table_layer_suppl}, the transferred results using input features better preserve the reference appearance compared to those using output features.

\section{Ablation study for time steps} \label{app:ablation_step}
\input{sources/suppl/abl_step}
Our method measures dense correspondence at timestep 92. 
Because our method performs sparse key-point matching in the early steps and dense matching in the later steps. 
As shown in \tref{tab:quan_table_dense_suppl}, the flow map distance is lower in the later steps compared to the early steps. 
It demonstrates that dense correspondence is more effectively measured in the later steps than in the early ones.

\section{Quantitative ablation results for each component} \label{app:quan_ablation}
\input{sources/ablation_table}
We add the below table to provide quantitative ablation for \fref{app:ablation}.
Feature rearrange is our core component and mainly helps structure $S_\text{miou}$). AdaIN adjusts color distribution ($A_\text{hist}$).

\section{Additional examples on ablation study} \label{app:ablation}
\input{sources/suppl/more_abl}
We present additional samples from the ablation study analyzing the effects of each component of our model in \fref{fig:more_ablation}. 

\section{Ablation study for object mask} \label{app:ablation_mask}
\input{sources/suppl/mask_setting}
\input{sources/suppl/mask}

\input{sources/suppl/mask_fig}
This section evaluates the role and effectiveness of object masks in appearance transfer tasks. 
\tref{tab:mask_setting} summarizes the approaches for obtaining object masks adopted by Ours and each baseline. 
Ours, DiffEditor, and ZeST utilize object masks during the transfer process, while other competitors do not incorporate object masks in their transfer processes.

To analyze the impact of object masks, we conduct experiments with our method without using an object mask. 
As shown in \tref{tab:more_mask}, the performance of Ours \textit{w/o mask} decreases in structure preservation compared to ZeST and DiffEditor, which use object masks. 
This result demonstrates that object masks are effective in maintaining the structure of the target object. 
Among competitors that do not use object masks, Ours \textit{w/o mask} achieves the best structure preservation. 
Regarding appearance similarity, our model maintains strong performance even without a mask, owing to its semantic matching capability during the transfer process.

\fref{fig:mask_fig} illustrates the appearance transfer results without using an object mask. Without an object mask, the background of the target image is not preserved after the transfer. This observation highlights that object masks ensure object-aware appearance transfer. Competitors that do not use object masks, such as Cross-Image, DiffEditor, DiffuseIT, SpliceViT, Swapping AE, and IP-Adapter, fail to preserve the background.

\section{Ablation study for matching rule}
\label{app:ablation_matching_rule}
\input{sources/suppl/aggregation}

\textbf{Rationale:} As we aim to transfer appearances according to semantic matches (e.g., beak-to-beak), it is natural to employ one-to-one winner-takes-all matches rather than softmax aggregation.

In Case 1, implicit alignments like softmax aggregation fail to preserve reference feature values. 
And in Case 2, the injection based on the matching between the query and key with the attention mechanism also produces similar failure results.
There are no scenarios where one-to-many or many-to-one matching outperforms one-to-one.
Features from similar regions inherently share similar values, so there are no cases where top-1 similarity is incorrect while top-2 to N is correct.
If cosine similarity fails in one-to-one matching, cosine similarity-based attention mechanisms would also fail.

\section{User study} \label{app:user_study}

\input{sources/user_study}
We conducted a user study with 53 participants, evaluating 15 randomly selected samples for appearance similarity (${U_\text{app}}$) and structure preservation (${U_\text{str}}$).

\section{Additional qualitative results} \label{app:qual}
\input{sources/suppl/more_result1}
\input{sources/suppl/more_result2}
\input{sources/suppl/more_result3}
\input{sources/suppl/more_result4}
\input{sources/suppl/more_multi}

In \fref{fig:more_result1} and \fref{fig:more_result2}, we provide more additional qualitative comparisons with competitors. 
In particular, \fref{fig:more_result1} illustrates the results when the reference and target are aligned but the reference object has complex patterns, as well as when the reference and target are unaligned.
And \fref{fig:more_result3} and \fref{fig:more_result4} showcase our transferred results across various domains.
Additionally, \fref{fig:more_multi} shows the results of appearance transfer from each object from two different reference images to multiple objects in a single target image. 
Each appearance transfer process occurs simultaneously.
\end{appendix}

%% file: sources/suppl/augment_data.tex
\begin{figure}[t]\centering
    \includegraphics[width=\linewidth]{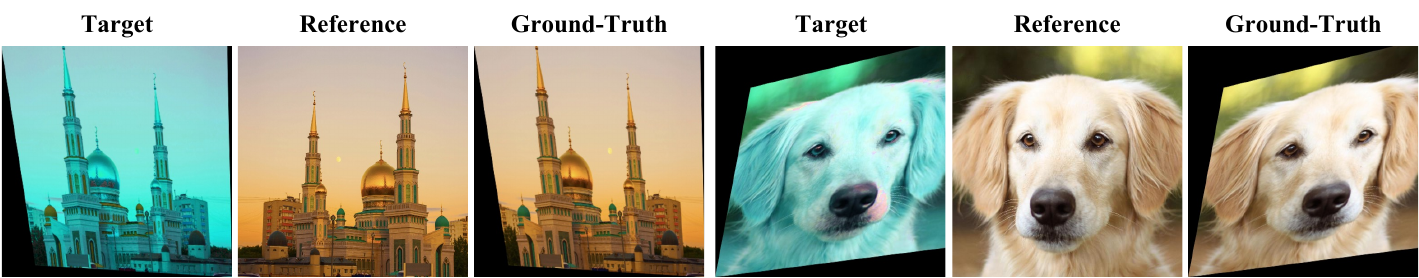}
    \caption{Examples of building and AFHQ for ${I_\text{hist}}$.}
    \label{fig:augment_data}
\end{figure}

%% file: sources/suppl/more_layer.tex
\begin{figure}[t]\centering
    \includegraphics[width=\linewidth]{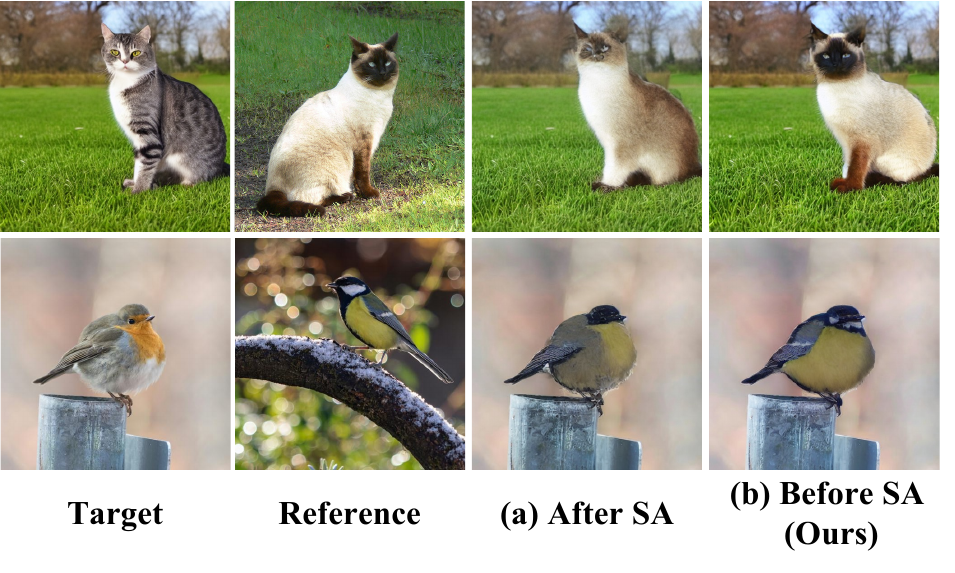}
    \caption{\textbf{Qualitative comparison between the results transferred using features after the self-attention layer and ours.} (a) Results transferred using features after the self-attention layer. (b) Results transferred using features before the self-attention layer (ours). (a) shows mismatched semantic correspondence, while (b) demonstrates accurate semantic correspondence.}
    \label{fig:more_layer}
\end{figure}

%% file: sources/suppl/abl_layer.tex
\begin{table}[]
\centering
\resizebox{0.8\linewidth}{!}{
\tiny  
\begin{tabular}{c|cc}
\specialrule{.05em}{1em}{0em} 
\hline
\multicolumn{1}{c|}{Metrics} &
\multicolumn{2}{c}{${A_\text{hist}} \downarrow$} \\
\hline 
\multicolumn{1}{c|}{Dataset} &
\multicolumn{1}{c}{Building} &
\multicolumn{1}{c}{AFHQ} \\
\hline
After SA            & 0.478 & 0.581 \\ 
Before SA(Ours)      & \textbf{0.469} & \textbf{0.577} \\ 
\hline
\end{tabular}
}
\caption{\textbf{Comparison of appearance similarity on different feature positions.} 
For all datasets, the appearance similarity of transferred results using features before the self-attention layer shows a lower $I_\text{hist}$ compared to those transferred using features after the self-attention layer. We mark the best score in bold.}
\label{tab:quan_table_layer_suppl}
\end{table}

%% file: sources/suppl/abl_step.tex
\begin{table}[]
\centering
\resizebox{\linewidth}{!}{
\scriptsize  
\begin{tabular}{c|ccc}
\specialrule{.05em}{1em}{0em} 
\hline
\multicolumn{1}{c|}{} & \multicolumn{3}{c}{${D_\text{flow}} \downarrow$} \\ \hline
\multicolumn{1}{c|}{\diagbox{Time Step}{Dataset}}& \multicolumn{1}{c}{FG3D CAR
} & \multicolumn{1}{c}{JODS} & \multicolumn{1}{c}{PASCAL} \\ \hline
62  & 10.75  & 32.86 & 28.37\\ 
77  & 9.71 & 30.16 & 24.72 \\
92(Ours) & \textbf{9.43} & \textbf{28.75} & \textbf{21.83} \\
\hline
\end{tabular}
}
\caption{\textbf{Comparison of dense correspondence on different time steps.} For all datasets, the dense correspondence measured at later time step shows a lower flow map distance compared to that measured at mid time step. We
mark the best score in bold.}
\label{tab:quan_table_dense_suppl}
\end{table}

%% file: sources/ablation_table.tex
\begin{table}[t]
\centering
\resizebox{\linewidth}{!}{
\begin{tabular}{l|ll|ll}
\specialrule{.05em}{1em}{0em} 
\hline
\multicolumn{1}{c|}{Method} 
& \multicolumn{2}{c|}{${S_\text{miou}} \uparrow$} 
& \multicolumn{2}{c|}{${A_\text{hist}} \downarrow$}\\ 
\hline
\multicolumn{1}{c|}{Component \textbackslash Dataset} 
& \multicolumn{1}{c}{Building}
& \multicolumn{1}{c|}{AFHQ} 
& \multicolumn{1}{c}{Building} 
& \multicolumn{1}{c|}{AFHQ} \\ 
\hline
Baseline(KV injection)   & 0.833 & 0.926 & 0.495 & 0.603 \\ 
+Feature rearrange     & 0.942 \textbf{(+0.109)} & 0.968 \textbf{(+0.038)} & 0.484 (-0.009) & 0.582 \textbf{(-0.021)} \\
+ AdaIN(Ours)   & 0.939 (-0.003) & 0.972 (+0.004) & 0.469 \textbf{(-0.015)} & 0.577 (-0.005) \\
\hline
\end{tabular}
}
\caption{\textbf{Quantitative ablation results.} We mark the greatest difference in scores between the components in bold.}
\label{tab:ablation_table}
\end{table}

%% file: sources/suppl/more_abl.tex
\begin{figure}[t]\centering
    \includegraphics[width=\linewidth]{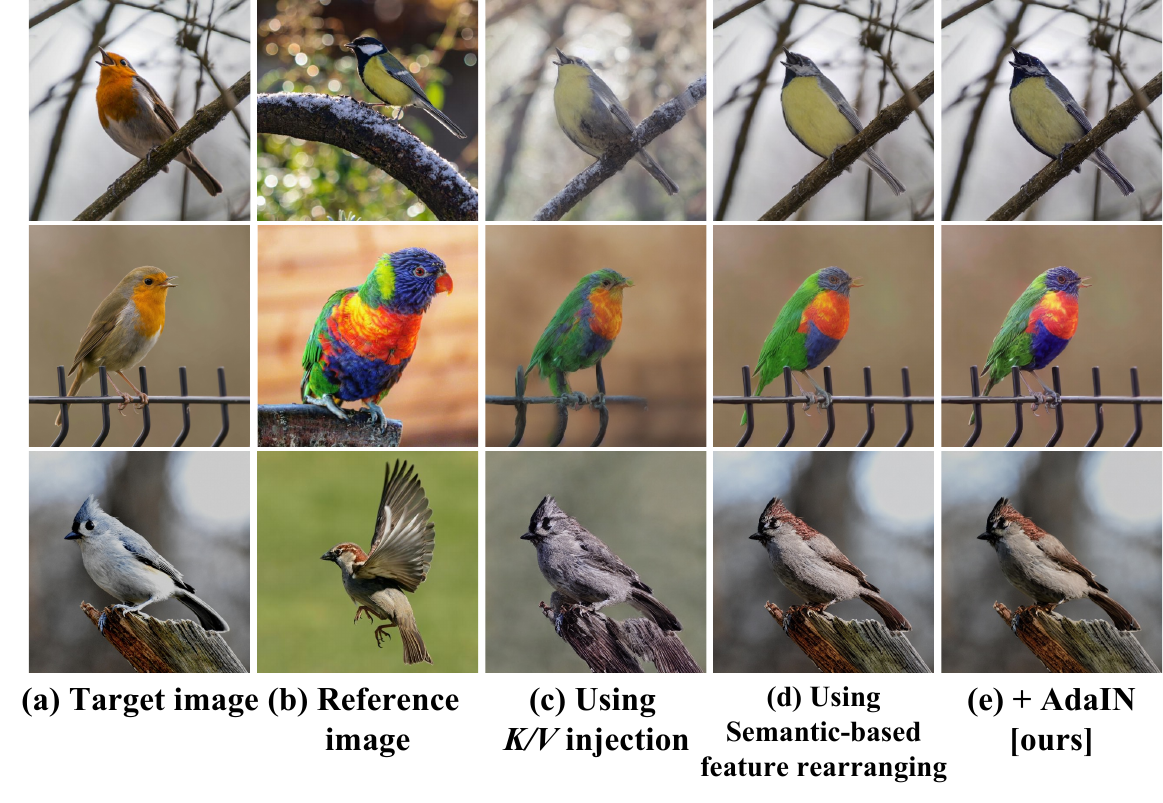}
    \caption{Additional samples of ablation study.}
    \label{fig:more_ablation}
\end{figure}

%% file: sources/suppl/mask_setting.tex
\begin{table}[]
\resizebox{\linewidth}{!}{

\begin{tabular}{c|cccc}
\specialrule{.05em}{1em}{0em} 
\hline
         & Ours   & DiffEditor   & ZeST  & Others \\ \hline
For mask & SAM-HQ \cite{samhq} & EfficientSAM \cite{efficientsam} & Rembg \cite{rembg} & X   \\ \hline
\end{tabular}
}

\caption{\textbf{Approaches for obtaining object masks.} The table compares the approaches used to obtain object masks in Ours, DiffEditor, and ZeST. Others refer to other baselines, including Cross-Image, DiffEditor, DiffuseIT, SpliceViT, Swapping AE, and IP-Adapter. These baselines do not utilize object masks.}
\label{tab:mask_setting}
\vspace{-3mm}
\end{table}

%% file: sources/suppl/mask.tex
\begin{table}[]
\centering
\resizebox{\linewidth}{!}{
\scriptsize  
\begin{tabular}{c|cc|cc}
\specialrule{.05em}{1em}{0em} 
\hline
\multicolumn{1}{c|}{Metrics} & \multicolumn{2}{c|}{${A_\text{hist}} \downarrow$} & \multicolumn{2}{c|}{${S_\text{miou}} \uparrow$} \\ \hline
\multicolumn{1}{c|}{Dataset} & \multicolumn{1}{c}{Building} & \multicolumn{1}{c|}{AFHQ} & \multicolumn{1}{c}{Building} & \multicolumn{1}{c}{AFHQ} \\ \hline
Ours *         & \cellcolor[HTML]{FFFFC7}0.469  & \cellcolor[HTML]{FFCCC9}0.577 & \cellcolor[HTML]{FFCCC9}0.939 & \cellcolor[HTML]{FFFFC7}0.972 \\ 
Ours \textit{w/o mask}    & \cellcolor[HTML]{FFCCC9}0.467  & \cellcolor[HTML]{FFFFC7}0.579 & 0.858 & 0.943 \\ 
Cross-Image      & 0.491  & 0.608 & 0.758 & 0.915 \\
DiffEditor *      & 0.478  & 0.608 & 0.863 & 0.943 \\
DiffuseIT        & 0.477  & 0.607 & 0.855 & 0.951 \\
Splice ViT       & 0.472  & 0.580 & 0.842 & 0.943 \\
Swapping AE      & 0.481  & 0.629 & 0.821 & 0.942 \\ 
IP-Adapter       & 0.487  & 0.616 & 0.642 & 0.950 \\ 
ZeST *          & 0.497  & 0.602 & \cellcolor[HTML]{FFFFC7}0.925 & \cellcolor[HTML]{FFCCC9}0.980 \\ 
\hline
\end{tabular}
}
\caption{\textbf{Comparison of appearance similarity and structure preservation for our model without a mask.} Ours \textit{w/o mask} refers to our method without using an object mask. * indicates a model using a mask. We mark the best score in red and the second-best score in yellow.}
\label{tab:more_mask}
\end{table}

%% file: sources/suppl/mask_fig.tex
\begin{figure}[t]\centering
    \includegraphics[width=\linewidth]{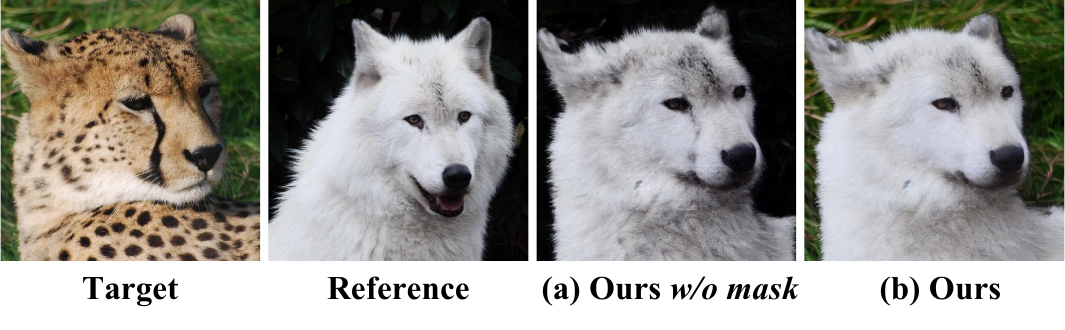}
    \caption{\textbf{Qualitative comparison between our model without an object mask and the our model.} (a), which does not apply the object mask during transfer, fails to preserve the background of the target image, whereas (b), with the mask applied, successfully retains the background.}
    \label{fig:mask_fig}
\end{figure}

%% file: sources/suppl/aggregation.tex
\begin{figure}[t]\centering
    \includegraphics[width=\linewidth]{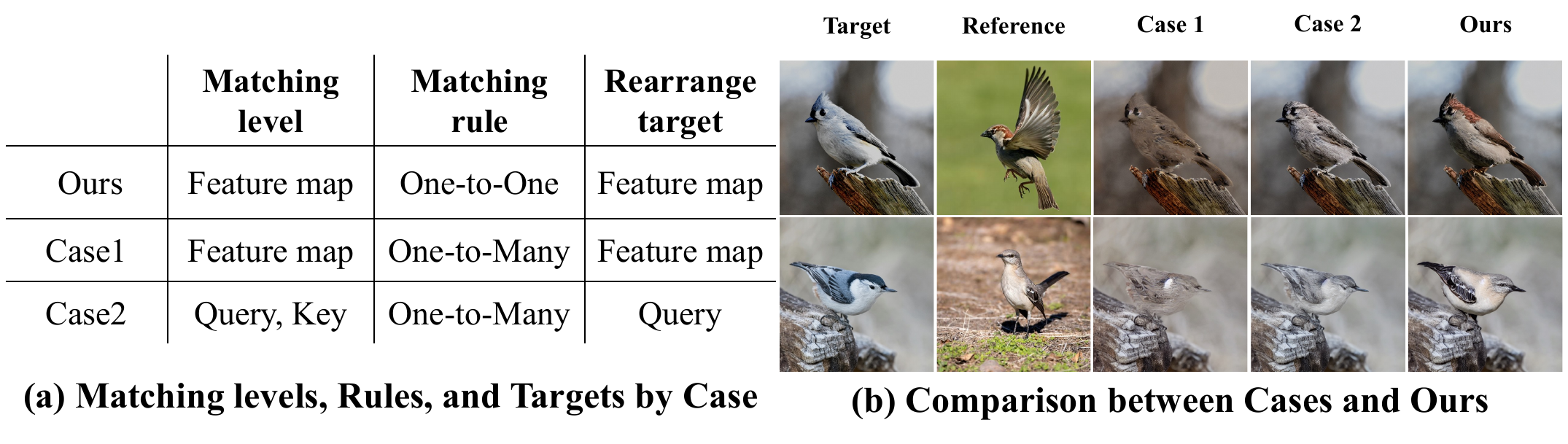}
    \caption{Comparison of matching component combinations.}
    \label{fig:aggregation}
\end{figure}

%% file: sources/user_study.tex
\begin{table}[h]
\centering
\resizebox{1\linewidth}{!}{
\begin{tabular}{c|ccccccc}
\specialrule{.05em}{1em}{0em} 
\hline
\multicolumn{1}{c|}{Method} &
\multicolumn{1}{c}{Ours} & 
\multicolumn{1}{c}{Cross-Image} &
\multicolumn{1}{c}{Diffeditor} &
\multicolumn{1}{c}{DiffuseIT} &
\multicolumn{1}{c}{Splice VIT} &
\multicolumn{1}{c}{IP-Adapter} &
\multicolumn{1}{c}{Zest} \\
\hline 
${U_\text{app}} $ & \textbf{0.661} & 0.059 & 0.033 & 0.026 & 0.096 & 0.062 & 0.064 \\ 
${U_\text{str}} $ & \textbf{0.462} & 0.014 & 0.042 & 0.121 & 0.030 & 0.062 & 0.276 \\
\hline
\end{tabular}
}
\caption{User study results. The bold is the best score.}
\label{tab:user_study}
\end{table}

%% file: sources/suppl/more_result1.tex
\begin{figure*}[t]\centering
    \includegraphics[width=\linewidth]{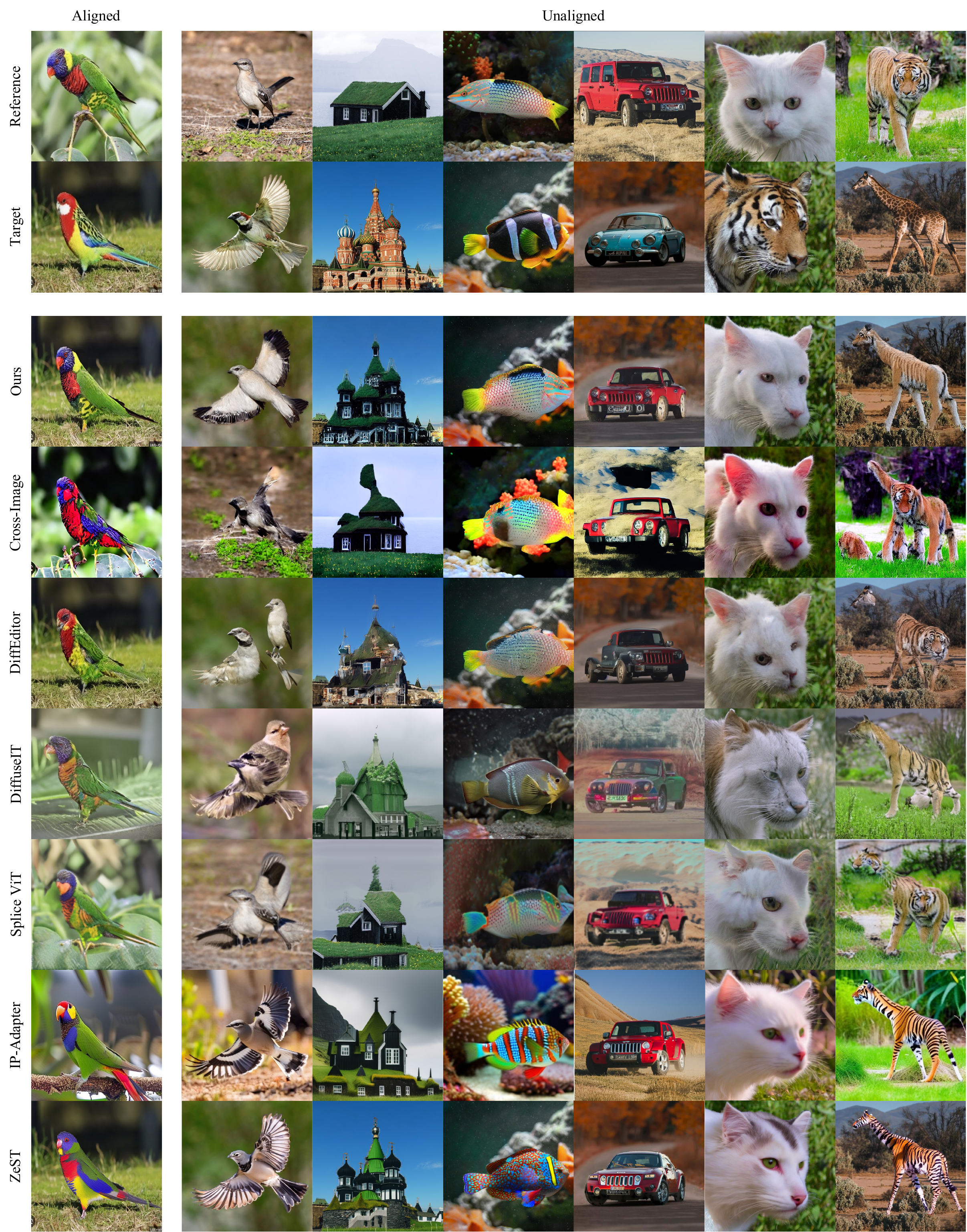}
    \caption{Our results on samples where the reference and target are aligned but the reference has complex patterns, as well as on various samples where the reference and target are misaligned.}
    \label{fig:more_result1}
\end{figure*}

%% file: sources/suppl/more_result2.tex
\begin{figure*}[t]\centering
    \includegraphics[width=\linewidth]{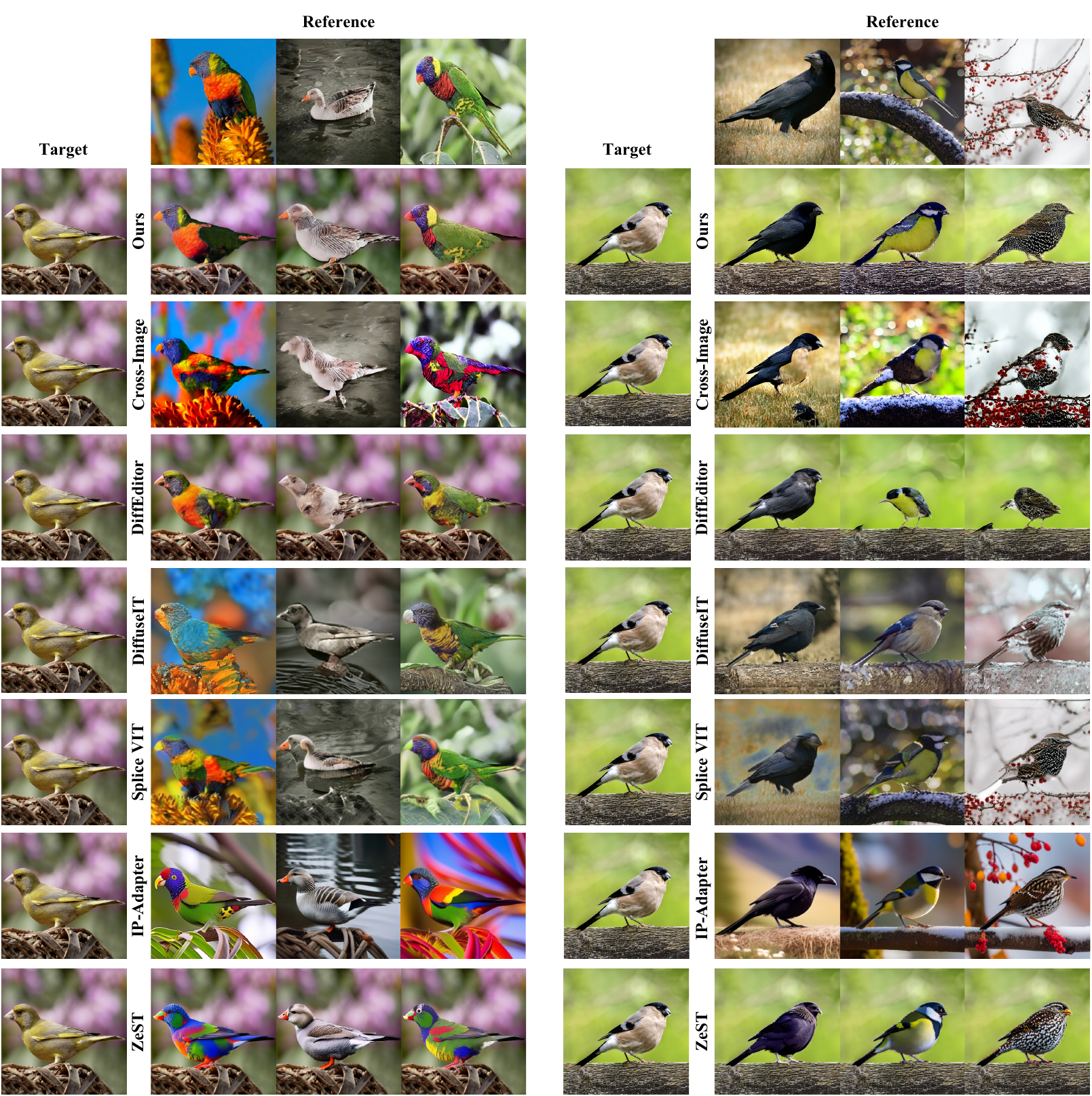}
    \caption{Qualitative comparison of appearance transfer for bird samples.}
    \label{fig:more_result2}
\end{figure*}

%% file: sources/suppl/more_result3.tex
\begin{figure*}[t]\centering
    \includegraphics[width=0.75\linewidth]{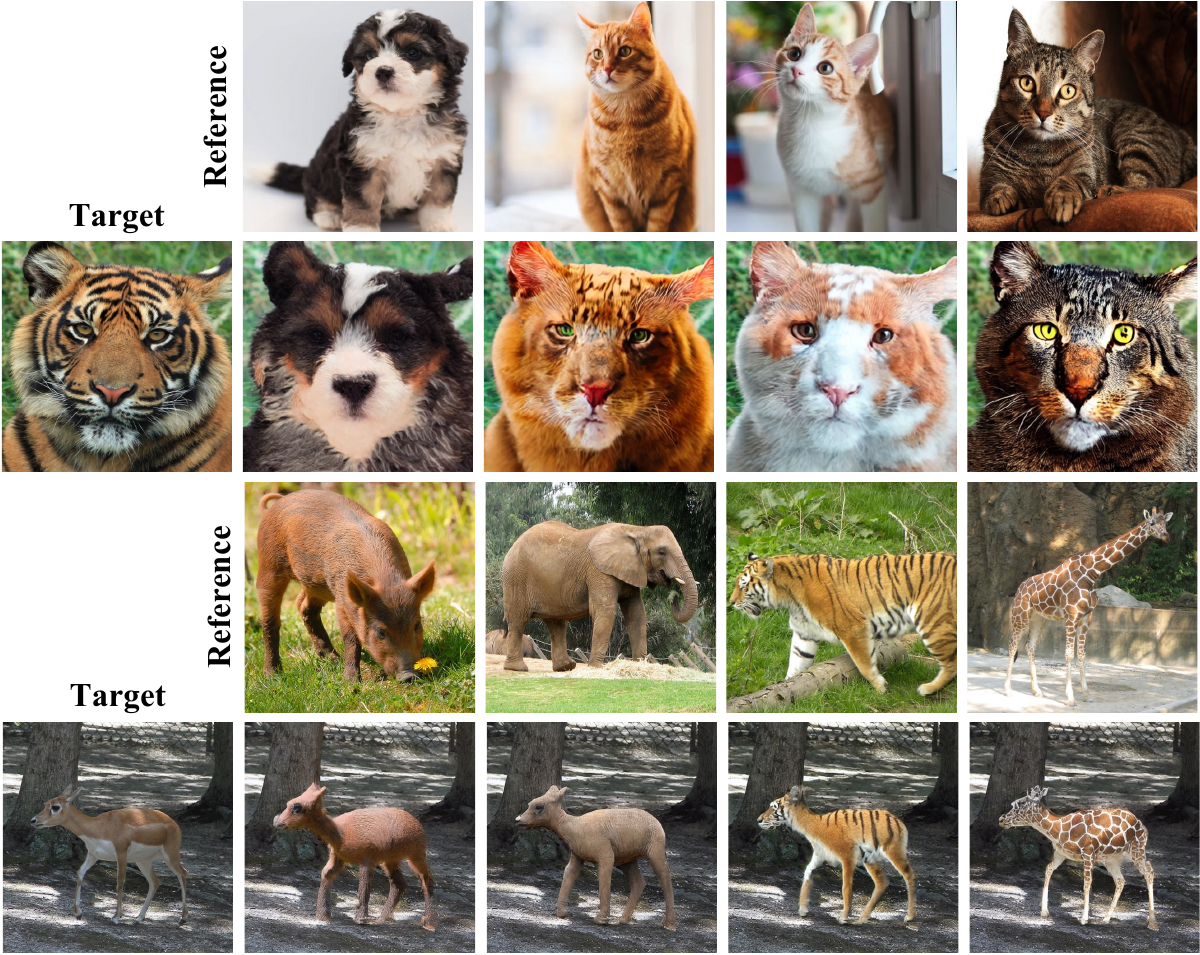}
    \caption{Our results on samples where the reference and target differ in size or are misaligned.}
    \label{fig:more_result3}
\end{figure*}

%% file: sources/suppl/more_result4.tex
\begin{figure*}[t]\centering
    \includegraphics[width=0.75\linewidth]{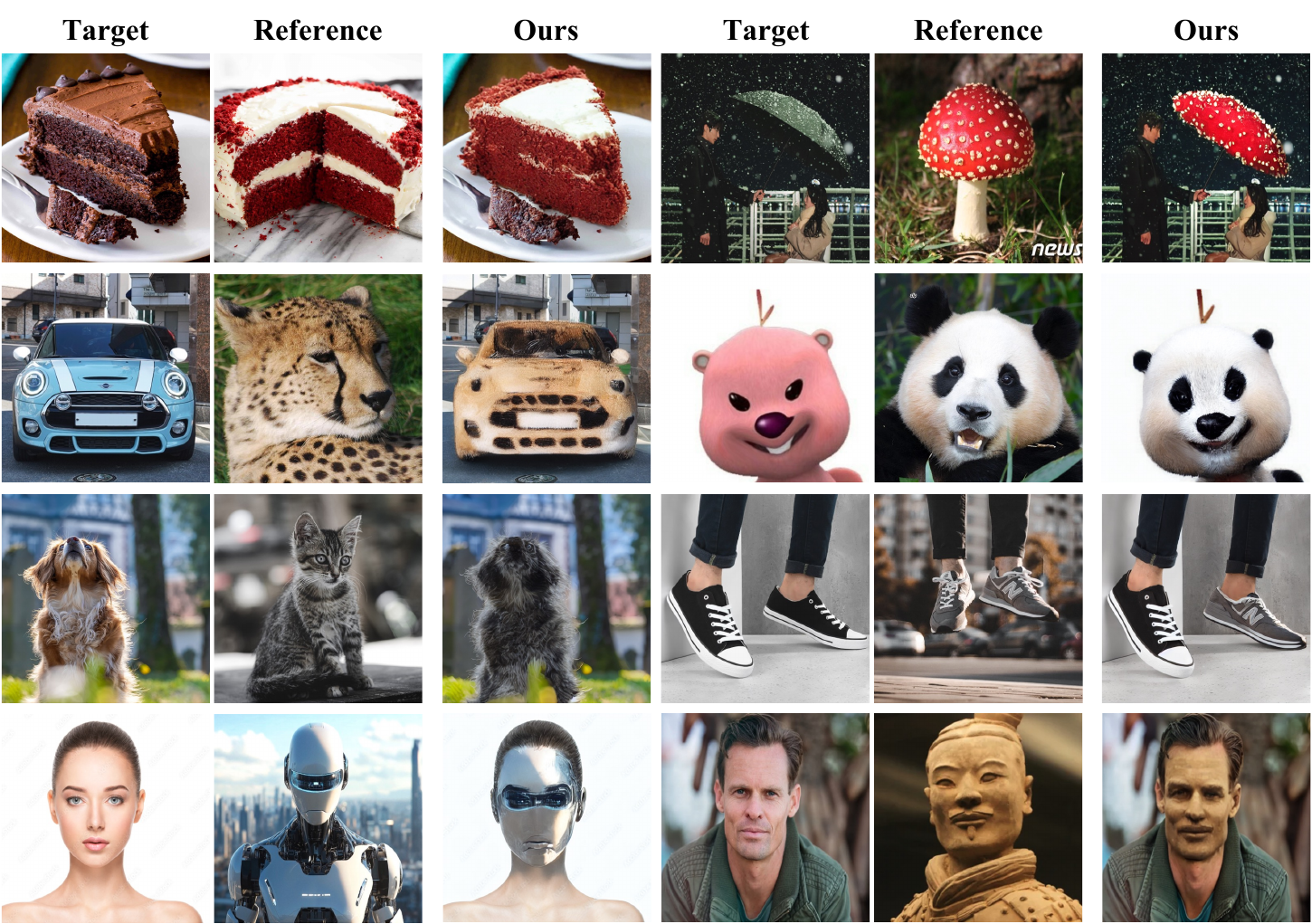}
    \caption{Our results of various domain.}
    \label{fig:more_result4}
\end{figure*}

%% file: sources/suppl/more_multi.tex
\begin{figure*}[t]\centering
    \includegraphics[width=0.75\linewidth]{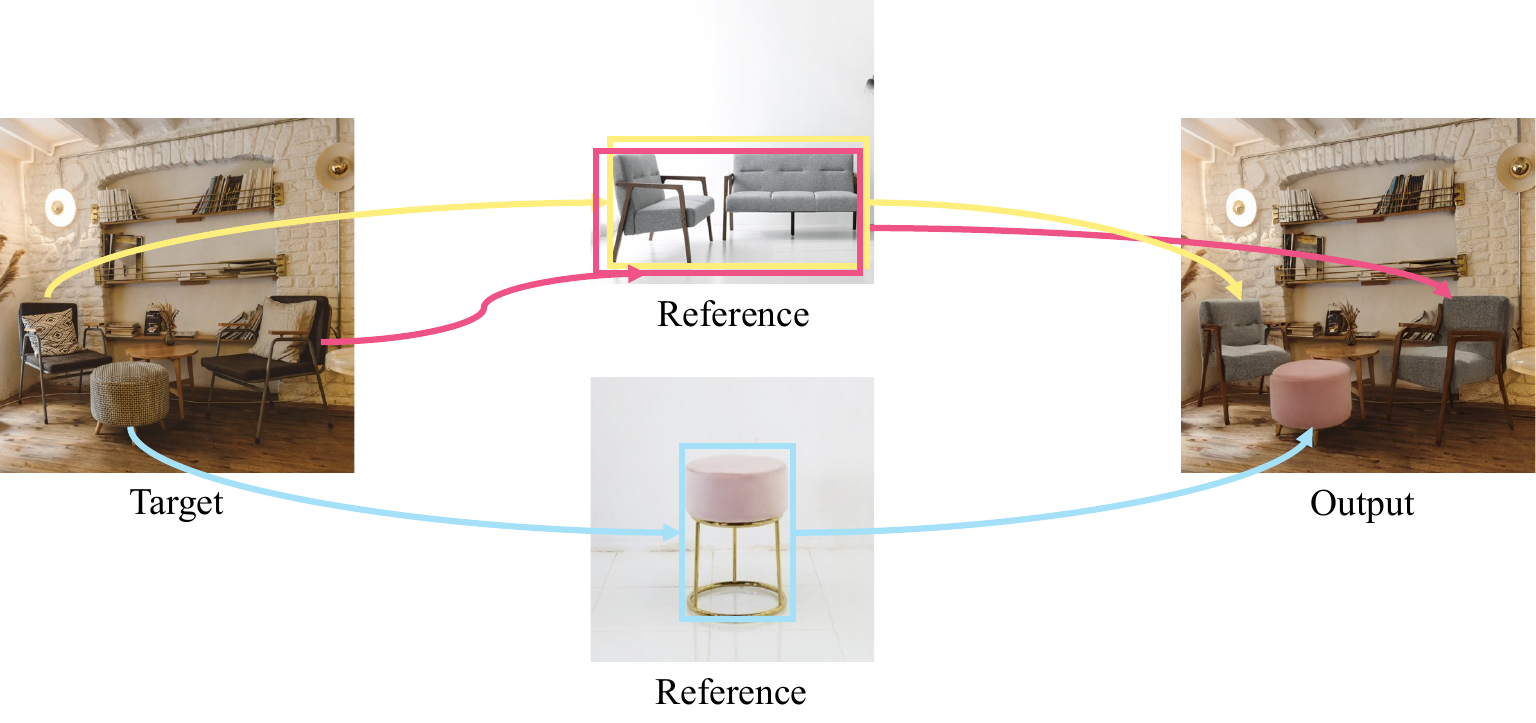}
    \includegraphics[width=0.75\linewidth]{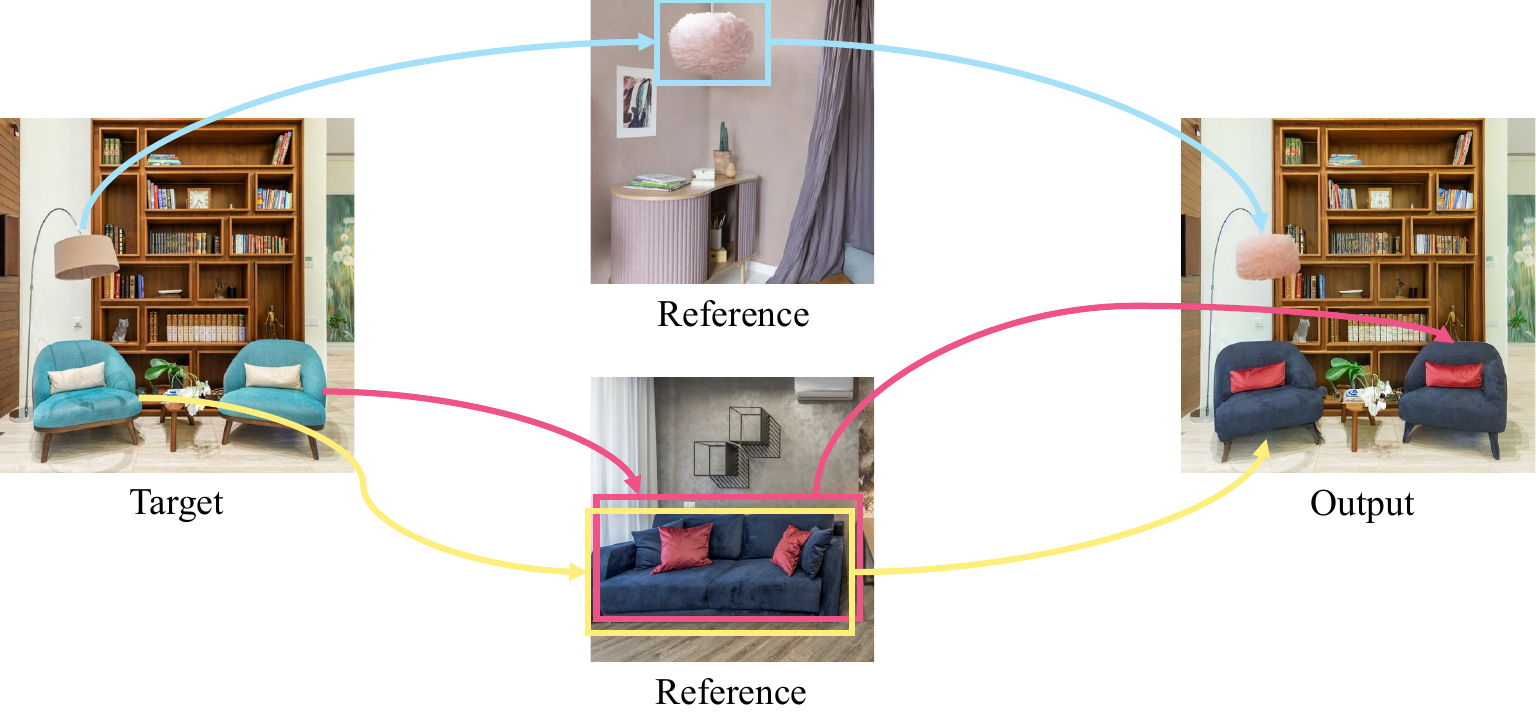}
    \caption{Results of appearance transfer between multiple objects.}
    \label{fig:more_multi}
\end{figure*}